\documentclass[10pt]{article} % For LaTeX2e
\usepackage[accepted]{tmlr}
\usepackage{amsmath,amsfonts,bm}

\def\eqref#1{equation~\ref{#1}}
\def\1{\bm{1}}

\DeclareMathAlphabet{\mathsfit}{\encodingdefault}{\sfdefault}{m}{sl}
\SetMathAlphabet{\mathsfit}{bold}{\encodingdefault}{\sfdefault}{bx}{n}

\usepackage{hyperref}
\usepackage{url}

\usepackage[utf8]{inputenc}
\usepackage[T1]{fontenc}
\usepackage{algorithm}
\usepackage{algorithmic}
\usepackage{graphicx}
\usepackage{booktabs}
\usepackage{multirow}
\usepackage{amsmath}
\usepackage{amsfonts}
\usepackage{enumitem}
\usepackage{url} 
\usepackage{xspace}
\usepackage[table]{xcolor}
\usepackage{hyperref}
\usepackage{epigraph}
\usepackage{amsthm}
\usepackage{amssymb}
\usepackage{arydshln}
\usepackage{bbm}
\usepackage{wrapfig}
\usepackage{subcaption}

\theoremstyle{plain}

\theoremstyle{definition}

\theoremstyle{remark}

\definecolor{langlightblue}{rgb}{0.3, 0.65, 1}
\definecolor{langblue}{rgb}{0, 0.4, 0.8}
\definecolor{langmildblue}{rgb}{0.0, 0.45, 0.73}
\definecolor{langdarkblue}{rgb}{0.0, 0.0, 0.61}
\definecolor{langred}{rgb}{0.81, 0.09, 0.13}
\definecolor{langgreen}{rgb}{0.0, 0.6, 0.3}
\definecolor{bingpink}{rgb}{1.0, 0.41, 0.71}
\hypersetup{
    colorlinks=true,
    linkcolor=langdarkblue,
    linkcolor=langblue,
    citecolor=langlightblue,
    urlcolor=langdarkblue,
}

\makeatletter
\AtBeginDocument{%
  \let\oldref\ref
  \renewcommand{\ref}[1]{%
    \hyperref[{#1}]{\underline{\oldref{#1}}}%
  }%
}
\makeatother

\usepackage{titletoc}
\newcommand\DoToC{%
  \startcontents
  \printcontents{}{1}{\textbf{\large Contents of Appendix}\vskip3pt\hrule\vskip5pt}
  \vskip3pt\hrule\vskip5pt
}
\title{Fortune: Formula-Driven Reinforcement Learning for LLM Reasoning over Complex Tables}

\author{\name Lang Cao \email langcao2@illinois.edu \\
\addr University of Illinois Urbana-Champaign
\AND
\name Jingxian Xu \email xujx@nbjl.nankai.edu.cn \\
\addr Nankai University
\AND
\name Hanbing Liu \email liuhb24@mails.tsinghua.edu.cn \\
\addr Tsinghua University
\AND
\name Jinyu Wang \email 202100150055@mail.sdu.edu.cn \\
\addr Shandong University
\AND
\name Mengyu Zhou \email zhoumengyu.zmy@alibaba-inc.com \\
\addr Qwen Large Model Application Team, Alibaba
\AND
\name Haoyu Dong \email hadong@microsoft.com \\
\addr Microsoft Research
\AND
\name Shi Han \email shihan@microsoft.com \\
\addr Microsoft Research
\AND
\name Dongmei Zhang \email dongmeiz@microsoft.com\\
\addr Microsoft Research
}

\def\month{11}  % Insert correct month for camera-ready version
\def\year{2026} % Insert correct year for camera-ready version
\def\openreview{\url{https://openreview.net/forum?id=1ajZG0sKKz}} % Insert correct link to OpenReview for camera-ready version

\begin{document}

\maketitle

% keywords: Table Understanding, Formula Learning, Symbolic Reasoning, Large Language Models
% TL;DR: Formula Tuning (Fortune) is a reinforcement learning approach that enables language models to perform symbolic table reasoning by deriving executable spreadsheet formulas.

\begin{abstract}
Tables are a fundamental medium for organizing and analyzing data, making table reasoning an essential capability for intelligent systems. Although large language models (LLMs) exhibit strong general reasoning abilities, they still struggle with accurate numerical reasoning over tabular data, particularly in complex table settings that require more than simple relational lookup. Spreadsheet formulas provide a powerful and expressive interface for executable symbolic operations, enabling sophisticated reasoning patterns that remain largely underexplored by existing LLMs. In this paper, we introduce \textit{Formula Tuning (Fortune)}, a formula-driven reinforcement learning (RL) framework that trains LLMs to generate executable spreadsheet formulas for question answering over tabular data, using execution success and answer correctness as reward signals. Building upon \textit{Fortune}, we train \textit{Formula-R1} and evaluate its performance across seven table reasoning benchmarks, demonstrating competitive performance, particularly on tasks involving complex tables and multi-step numerical computation. Furthermore, \textit{Formula-R1-Plus}, which integrates textual and formula-based reasoning, achieves further performance improvements. Beyond these empirical results, we provide insights into the roles of reinforcement learning and executable formula representations in table reasoning, highlighting the broader potential of symbolic RL to enhance the reasoning capabilities of LLMs.
\end{abstract}

% paper takeaways:

% motivation:
%   better table understanding
%   utilization of formula but lack of data
%   better symbolic reasoning

% methods:
%   reinforcement learning

% others:
%   general table understanding / general two-dimensional data

\section{Introduction}

Tables are a common and practical data structure in daily life, playing a central role in data collection, representation, and analysis \citep{he2023text2analysisbenchmarktablequestion, yi2025tablepilotrecommendinghumanpreferredtabular}. Recent advances in large language models (LLMs) \citep{gunasekar2023textbooksneed, openai2024gpt4technicalreport, touvron2023llamaopenefficientfoundation} have brought impressive performance across a wide range of natural language processing tasks, including language understanding \citep{minaee2024largelanguagemodelssurvey, zhu-etal-2024-large} and general reasoning \citep{plaat2024reasoninglargelanguagemodels}. Naturally, LLMs have also been applied to tabular data understanding and reasoning \citep{fang2024largelanguagemodelsllmstabular, zhang2024surveytablereasoninglarge, cao2025tablemasterrecipeadvancetable}.

However, reasoning over tabular data remains a key challenge for LLMs, primarily because of the numerical nature and intricate structure of tables, which pose significant obstacles to understanding both content and layout \citep{cao2025tablemasterrecipeadvancetable}. Besides, some findings \citep{yan2025phdlevelllmstrulygrasp} suggest current LLMs rely on pattern memorization over genuine rule learning, often leading to incorrect mathematical computations during traditional chain-of-thought reasoning, also referred to as textual reasoning. Some approaches incorporate symbolic reasoning methods \citep{chen2023programthoughtspromptingdisentangling, gao2023palprogramaidedlanguagemodels}, where LLMs first generate symbolic representations, such as programs used as symbolic tools, and then execute them to obtain final results. While this improves arithmetic accuracy, such methods often struggle to generalize due to limited symbolic reasoning or program generation capabilities. Instead of truly understanding the context and generating problem-solving programs, LLMs often fall back on memorized code snippets from pretraining \citep{yang-etal-2024-llms}. Since not all complex symbolic patterns can be memorized, and given that high-quality code supervision is scarce \citep{bi2023programofthoughtsworkreasoning}, more effective strategies for symbolic table reasoning are needed.

Recent progress in reinforcement learning (RL) for LLMs shows promise for overcoming such limitations. For example, DeepSeek-R1 \citep{deepseekai2025deepseekr1incentivizingreasoningcapability} improves mathematical reasoning in LLMs via rule-based RL rewards, without relying on step-by-step annotations. DeepRetrieval \citep{jiang2025deepretrievalhackingrealsearch} uses retrieval metrics as RL rewards to train models to reason over queries that maximize real-world retrieval performance across search engines and databases. DeepCoder \citep{deepcoder2025} also demonstrates the effectiveness of RL for code reasoning and generation. These works collectively demonstrate the promise of RL in enabling LMs to perform robust symbolic reasoning without explicit intermediate supervision.

% \footnote{\href{https://support.microsoft.com/en-us/office/overview-of-formulas-34519a4e-1e8d-4f4b-84d4-d642c4f63263}{https://support.microsoft.com/en-us/office/overview-of-formulas}}

When it comes to symbolic reasoning over tables, spreadsheet formulas \citep{microsoft_excel_formulas} is a powerful and versatile tool. In real-world scenarios, tabular data is often stored in spreadsheet formats (e.g., Microsoft Excel, Google Sheets) \citep{dong-etal-2024-encoding}, where each cell holds individual data values. The spreadsheet formulas embedded in these files act as lightweight, program-like constructs that enable users to compute, transform, and reason over data. Compared to structured interfaces such as SQL or Python/Pandas, which are typically restricted to relational or flat table schemas, spreadsheet formulas offer greater flexibility, as they can be applied to arbitrary two-dimensional tables without structural constraints \citep{wang2025generaltablequestionanswering}. Moreover, spreadsheet formulas are Turing complete \citep{smalley2023excel}, making them particularly well-suited as a medium for symbolic table reasoning.

We envision that training LLMs to understand and generate spreadsheet formulas can help them develop more robust and generalizable symbolic reasoning capabilities over tabular data. Here, symbolic reasoning refers to manipulating abstract symbols according to formal rules, while formula-based reasoning refers to expressing and executing such reasoning through spreadsheet formulas. However, current LLMs still struggle to produce accurate and reliable spreadsheet formulas, as highlighted by recent evaluations \citep{thorne2023experimentingchatgptspreadsheetformula}. At the same time, existing publicly available spreadsheet datasets tend to include relatively simple formulas \citep{cheng2021hitab}, rely on heuristic conversions from SQL \citep{zhao-etal-2024-nl2formula}, or synthesize formulas using LLMs in constrained question-answering settings \citep{wang2025generaltablequestionanswering}. These approaches fall short of capturing the complexity necessary for diverse symbolic reasoning tasks and real-world downstream applications. This limitation poses a significant barrier to effectively leveraging spreadsheet formulas for training LLMs in symbolic table reasoning.

To address these challenges, we propose \textit{\textbf{For}mula \textbf{Tun}ing} (\textbf{Fortune}), an RL framework designed to teach LLMs to perform symbolic reasoning over general tabular data through spreadsheet formulas. Specifically, our framework uses execution success and answer correctness from formula execution as reward signals to guide LLMs in deriving formulas through reasoning. This design reduces reliance on supervised formula annotations and enables LLMs to generate executable formulas that answer questions over tables more accurately. Extensive experiments on seven benchmark validate the effectiveness of \textit{Formula Tuning}, showing that RL is more effective than supervised fine-tuning (SFT) at enhancing the symbolic reasoning capabilities of LLMs. We also find that initializing RL with an SFT cold start \citep{deepseekai2025deepseekr1incentivizingreasoningcapability} provides a stronger foundation and improves the upper bound of RL performance, with SFT serving as a form of knowledge injection. Based on this framework, we train \textbf{Formula-R1}. Furthermore, we train both textual and symbolic reasoning components with RL in \textbf{Formula-R1-Plus}. By jointly leveraging both components during inference, our method achieves strong performance across multiple benchmarks, including 82.54\% on WikiTQ, 95.06\% on TabFact, 87.24\% on HiTab, 80.47\% on FinQA, and 93.20\% on AIT-QA, outperforming previous methods on all five publicly available benchmarks.

% state-of-the-art

In summary, this paper makes the following key contributions:
\begin{itemize}[leftmargin=*, itemsep=0pt, labelsep=5pt, topsep=0pt]
    \item We propose \textit{Formula Tuning} (\textbf{Fortune}), a reinforcement learning framework that enhances symbolic reasoning for table understanding by training large language models to generate executable spreadsheet formulas.
    \item We conduct extensive experiments on seven table understanding benchmarks to validate the effectiveness of \textit{Formula Tuning}, and we develop \textbf{Formula-R1} based on this training framework.
    \item We provide comprehensive empirical analyses, offering deeper insights into textual versus symbolic reasoning for table understanding, as well as supervised fine-tuning versus reinforcement learning for symbolic table reasoning.
\end{itemize}

\section{Related Work}

\noindent \textbf{Table Understanding and Reasoning.}
% finetuning
Many studies have explored fine-tuning language models (LMs) to improve their ability to understand and reason over tabular data. Building on the masked language modeling introduced by BERT \citep{devlin2019bertpretrainingdeepbidirectional}, models such as TaPas \citep{herzig2020tapas}, PaSTA \citep{gu2022pastatableoperationsawarefact}, and TUTA \citep{wang2021tuta} propose specialized pretraining strategies tailored for tables. TAPEX \citep{liu2022tapextablepretraininglearning} pretrains an encoder-decoder model as a neural SQL executor to better capture the semantics of table operations. Recent efforts, including TableLLaMA \citep{zhang2024tablellamaopenlargegeneralist} and TableGPT \citep{zha2023tablegptunifyingtablesnature}, build upon large decoder-only language models pretrained for general-purpose table understanding across a wide range of downstream tasks.

% zero-shot symbolic reasoning
Other studies focus on enabling LMs to better perform table-related tasks without fine-tuning. For example, Dater \citep{ye2023largelanguagemodelsversatile} proposes strategies for dynamically constructing sub-tables, modifying the input context to enhance comprehension. Chain-of-Table \citep{wang2024chainoftableevolvingtablesreasoning} models table reasoning as a sequence of transformations using predefined operations, gradually generating sub-tables to support complex multi-step inference. TableMaster \citep{cao2025tablemasterrecipeadvancetable} introduces a general framework for table understanding and underscores the importance of symbolic reasoning in handling complex scenarios. Given the structured and often numerical nature of tabular data, program-of-thought prompting \citep{chen2023programthoughtspromptingdisentangling} and other symbolic approaches \citep{cheng2023bindinglanguagemodelssymbolic, nahid-rafiei-2024-tabsqlify, mao2024potableprogrammingstandardlytablebased} have demonstrated strong effectiveness for table reasoning.

Some concurrent studies, including Table-R1 \citep{wu2026table}, Reasoning-Table \citep{lei2025reasoning}, and SQL-R1 \citep{ma2026sql}, also explore reinforcement learning with verifiable rewards (RLVR) for structured reasoning tasks. However, Table-R1 focuses on identifying relevant table regions, Reasoning-Table investigates general table reasoning, and SQL-R1 targets natural-language-to-SQL generation. In contrast, our work uses spreadsheet formulas as an executable reasoning representation and provides both theoretical analysis and empirical evidence demonstrating their advantages for complex numerical table reasoning.
\\
% ours - trained symbolic reasoning

\noindent \textbf{Formula Learning.}
A growing body of research has explored the potential of spreadsheet formulas as a powerful means to enhance table understanding. NL2Formula \citep{zhao-etal-2024-nl2formula} constructs a formula generation dataset by converting text-to-SQL tasks into spreadsheet formulas, enabling position-aware reasoning from natural language queries. ForTap \citep{cheng2022fortapusingformulasnumericalreasoningaware} leverages spreadsheet formulas as pretraining signals to enhance numerical reasoning. Auto-Formula \citep{chen2024auto} applies contrastive learning to transfer formulas from similar spreadsheets for formula recommendation. SpreadsheetCoder \citep{chen2021spreadsheetcoderformulapredictionsemistructured} formulates formula prediction as a program synthesis task, leveraging both headers and surrounding cell values for context. FLAME \citep{joshi2023flamesmalllanguagemodel} trains a small domain-specific model tailored for formula repair and completion. TabAF \citep{wang2025generaltablequestionanswering} jointly generates answers and formulas for table question answering, but relies on supervised fine-tuning over datasets generated by LLMs.
% Some studies \citep{singh2024empiricalstudyvalidatingsynthetic} show that validating synthetic natural language annotations can improve LLM fine-tuning for formula generation.
\\

\noindent \textbf{Reinforcement Learning for Language Models.}
In the era of large language models (LLMs), reinforcement learning (RL) \citep{Kaelbling1996ReinforcementLA} has gained significant traction as an effective framework for aligning models with human preferences. A prominent example is Reinforcement Learning from Human Feedback (RLHF) \citep{Christiano2017DeepRL, Stiennon2020LearningTS, Ouyang2022TrainingLM}, which leverages the Proximal Policy Optimization (PPO) algorithm \citep{Schulman2017ProximalPO} and human preference data to train a reward model that guides the fine-tuning of LLMs. In terms of application, DeepRetrieval \citep{jiang2025deepretrievalhackingrealsearch} applies RL to teach models how to reason about interacting with search engines for information retrieval, while DeepCoder \citep{deepcoder2025} uses RL for code reasoning and generation tasks. Rec-R1 \citep{lin2025recr1bridginggenerativelarge} also bridges LLMs and recommendation systems through RL.
\\

\begin{figure*}[t!]
    \centering
    \includegraphics[width=0.9\textwidth]{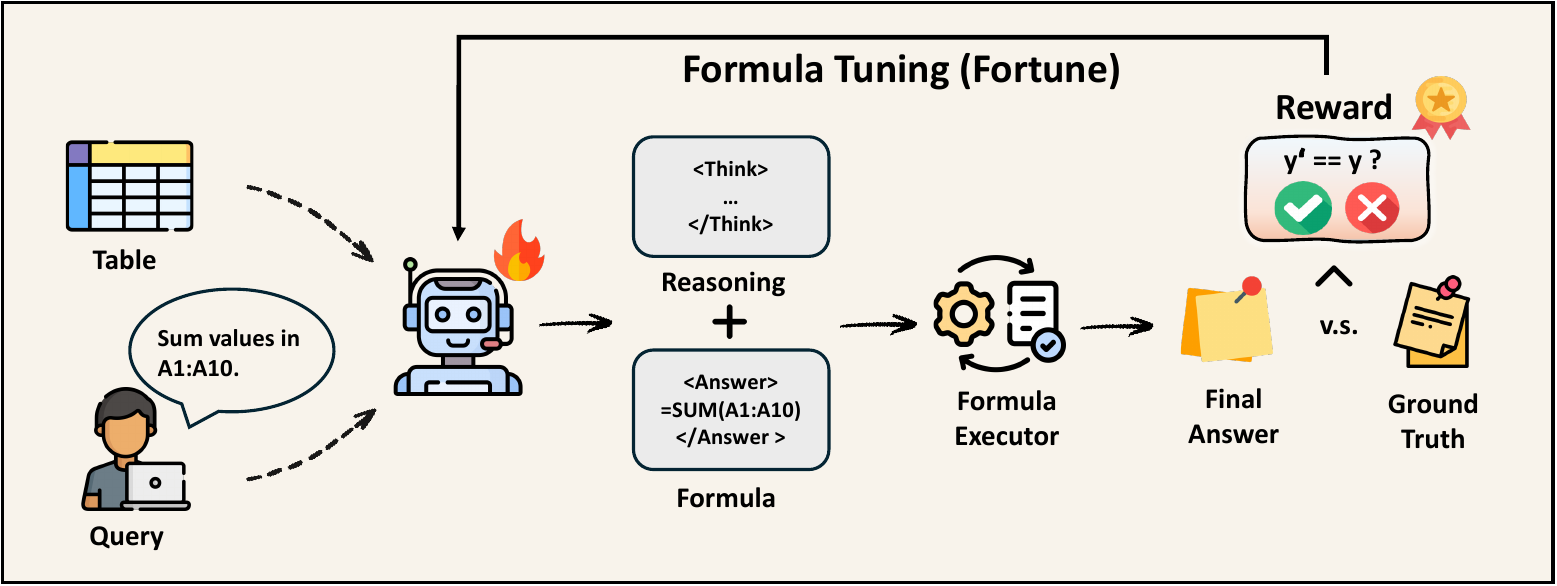}
    \caption{Overview of \textit{Formula Tuning} (Fortune).}
    \label{fig:framework}
    % Acknowledging Jingxian for the artwork A.
\end{figure*}

\section{Methodology}
% In this section, we present a theoretical analysis and discussion comparing textual versus symbolic reasoning in table understanding, as well as supervised fine-tuning (SFT) versus reinforcement learning (RL) in symbolic table reasoning (Figure~\ref{fig:method}). We then introduce our proposed training framework, \textit{Formula Tuning}. All notations are listed at Appendix~\ref{ap:notation}.
In this section, we introduce our proposed training framework, \textit{Formula Tuning}, illustrated in Figure~\ref{fig:framework}.
% A summary of all notations is provided in Appendix~\ref{ap:notation}.

\subsection{Task Formulation}
\textbf{Table Understanding with Language Models.}  
We consider a language model (LM) as a conditional generation policy \(\pi_{\theta}(a \mid s)\), where \(\theta\) denotes its parameters. The input \(s \in \mathcal{S}\) comprises a table \(\mathbb{T}\) and a natural-language query \(q\), i.e., \( s = (\mathbb{T}, q) \). The table \(\mathbb{T}\) is a two-dimensional grid of cells:
\begin{equation}
\mathbb{T}_{m \times n}
=
\begin{bmatrix}
C_{1,1} & C_{1,2} & \cdots \\ 
C_{2,1} & C_{i,j} & \cdots \\ 
\vdots  & \vdots  & \ddots \\ 
\end{bmatrix},
\end{equation}
where each \(C_{i,j}\) may contain a data value, structural information (e.g., a top header or a left header), or be empty. In practice, we linearize \(\mathbb{T}\) into a text sequence before feeding it to the LM. The LM then generates an output \(a \in \mathcal{A}\), which can be either a final textual answer or a spreadsheet formula \( f \) that produces the answer upon execution. Our goal is to optimize the parameters \(\theta\) that maximize the expected table‐understanding performance, measured by a reward function \(r(a \mid s)\). Formally,
\begin{equation}
\max_{\theta}\;
\mathbb{E}_{s \sim p(s),\,a \sim \pi_{\theta}(\cdot \mid s)}\bigl[r(a \mid s)\bigr],
\label{eq:objective}
\end{equation}
where \(p(s)\) denotes the empirical distribution over table–query pairs and \(r(a \mid s)\) evaluates the correctness of the final answer from the LM given the input.

\subsection{Formula Tuning}
\label{sec:fortune}

%-------------------------------------------------------
% Definition: Formula Tuning
%-------------------------------------------------------
\textbf{Definition.}  
\emph{Formula Tuning} is a reinforcement learning (RL) framework that defines spreadsheet formulas as an explicit symbolic reasoning space for table understanding. Specifically, we train a pretrained LM \(\pi_{\theta}\) to generate formulas \(f \in \mathcal{F}\), which are executed by a deterministic spreadsheet engine \(\operatorname{exec}(f, \mathbb{T})\). The reward is defined in two stages: execution success and answer correctness. First, the generated formula must be executable. If execution succeeds, the resulting answer \(a = \operatorname{exec}(f, \mathbb{T})\) is compared with the ground-truth answer \(a^{\star}(s)\). Formally, the model receives the following reward:

\begin{equation}
r(a \mid s) =
\begin{cases}
1,   & \text{if } a = a^{\star}(s), \\[0.3em]
0.2, & \text{if } a \ne a^{\star}(s) \text{ and } f \text{ is executable}, \\[0.3em]
0,   & \text{if } f \text{ is not executable}.
\end{cases}
\end{equation}

This reward function encourages the model to explore valid and executable formulas, even when they do not initially produce the correct answer, while assigning full credit only to formulas whose execution results exactly match the ground truth.

%-------------------------------------------------------
% Objective: RL optimization
%-------------------------------------------------------
\noindent
\textbf{Objective.}  
Formula Tuning maximizes the expected reward using RL algorithms, such as proximal policy optimization (PPO)~\citep{Schulman2017ProximalPO}, with the action space constrained to spreadsheet formulas:

\begin{equation}
\max_{\theta} \;
\mathbb{E}_{s \sim p(s),\, f \sim \pi_{\theta}(\cdot \mid s)}
\left[
r\left( \operatorname{exec}(f, \mathbb{T}) \mid s \right)
\right].
\end{equation}

%-------------------------------------------------------
% Workflow: Training process
%-------------------------------------------------------
\noindent
\textbf{Training Workflow.}
\begin{enumerate}[leftmargin=20pt]
\item \emph{Decoding:}\; The LM generates a chain of thought and samples candidate formulas \(f_1, f_2, \ldots\) from its current policy \(\pi_{\theta}\).
\item \emph{Execution:}\; Each formula \(f_k\) is executed to produce the corresponding answer \(a_k\).
\item \emph{Rewarding:}\; The environment returns a scalar reward \(r_k = r(a_k \mid s)\) based on the correctness and executability of the result.
\item \emph{Policy Update:}\; The LM parameters \(\theta\) are updated using a RL algorithm (e.g., PPO) based on the observed tuple \((s, f_k, r_k)\).
\end{enumerate}

\noindent
This framework enables the model to perform symbolic reasoning over complex tables with more accurate numerical computation. Additional methodological details are provided in Appendix~\ref{ap:discussion}.

% We further present a theoretical discussion comparing textual and symbolic reasoning in table reasoning, as well as SFT and RL in symbolic table reasoning, in Appendix~\ref{ap:theory}.

\section{Experiments}

In this section, we train \textit{Formula-R1} based on this framework. We present extensive experiments to empirically demonstrate the effectiveness of \textit{Formula Tuning}, and further provide additional analyses, discussion, and insights based on these observations.

\subsection{Settings}
We conduct experiments on seven diverse table understanding benchmarks, including WikiTQ~\citep{pasupat2015compositionalsemanticparsingsemistructured}, TabFact~\citep{chen2020tabfactlargescaledatasettablebased}, FinQA~\citep{chen-etal-2021-finqa}, HiTab~\citep{cheng2021hitab}, MultiHiertt~\citep{zhao-etal-2022-multihiertt}, AIT-QA~\citep{katsis2021aitqa}, and TableBench~\citep{wu2025tablebenchcomprehensivecomplexbenchmark}. These datasets differ in domain sources, table structure types, and question complexity, collectively covering the full spectrum of table understanding tasks. We do not compare against prior work on TableBench or MultiHiertt because these original studies do not report results on these datasets. For training, we merge the first five datasets into a single training corpus and train the model jointly on this combined set, then evaluate it separately on each dataset. Among these, AIT-QA and TableBench are treated as out-of-distribution (OOD) evaluation sets, while the rest are considered in-distribution (ID).
Our experiments primarily involve \textit{Llama-3.1$_{8\text{B}}$} and \textit{Qwen2.5-Coder$_{7\text{B}}$}. Since these base models are relatively small (7-8B parameters), their intrinsic ability to generate spreadsheet formulas is limited. To better equip them with symbolic table-reasoning knowledge, we apply a cold-start SFT stage \citep{deepseekai2025deepseekr1incentivizingreasoningcapability} before RL as initialization. The supervision data used in this stage is distilled from \textit{GPT-4o}. Following prior work~\citep{pasupat2015compositionalsemanticparsingsemistructured, cheng2021hitab}, we use exact match accuracy as our primary evaluation metric. The prompts used in our experiments are provided in Appendix~\ref{ap:prompt}. Detailed settings are provided in Appendix~\ref{ap:detail_setting}.

\begin{table*}[t!]
  \centering
  \caption{Performance comparison of different methods. Values in the table indicate accuracy (\%). Values marked with * indicate out-of-distribution results. `-' indicates results not reported in the related paper. For fine-tuning-based methods, the best performance in each column is highlighted in \textcolor{langdarkblue}{dark blue}, and the second-best in \textcolor{langmildblue}{light blue}.}
  \resizebox{1\textwidth}{!}{
    \begin{tabular}{llccccc}
    \toprule
    \textbf{Method} & \textbf{Backbone} 
      & \textbf{WikiTQ}\phantom{\textsuperscript{*}} 
      & \textbf{TabFact}\phantom{\textsuperscript{*}} 
      & \textbf{HiTab}\phantom{\textsuperscript{*}} 
      & \textbf{FinQA}\phantom{\textsuperscript{*}} 
      & \textbf{AIT-QA}\phantom{\textsuperscript{*}} \\
    \midrule
    \multicolumn{7}{l}{\textit{\textbf{Prompting-Based Methods}}} \\
    \quad Binder~\citep{cheng2023bindinglanguagemodelssymbolic} 
      & CodeX 
      & 64.60\phantom{\textsuperscript{*}} 
      & 85.10\phantom{\textsuperscript{*}} 
      & -\phantom{\textsuperscript{*}} 
      & -\phantom{\textsuperscript{*}} 
      & -\phantom{\textsuperscript{*}} \\
    \quad Dater~\citep{ye2023largelanguagemodelsversatile} 
      & CodeX 
      & 65.90\phantom{\textsuperscript{*}} 
      & 85.60\phantom{\textsuperscript{*}} 
      & -\phantom{\textsuperscript{*}} 
      & -\phantom{\textsuperscript{*}} 
      & -\phantom{\textsuperscript{*}} \\
    \quad API-Assisted~\citep{cao2025tablemasterrecipeadvancetable} 
      & CodeX 
      & 42.40\phantom{\textsuperscript{*}} 
      & -\phantom{\textsuperscript{*}} 
      & 69.30\phantom{\textsuperscript{*}} 
      & -\phantom{\textsuperscript{*}} 
      & -\phantom{\textsuperscript{*}} \\
    \quad ReAcTable~\citep{zhang2023reactableenhancingreacttable} 
      & CodeX 
      & 68.00\phantom{\textsuperscript{*}} 
      & 86.10\phantom{\textsuperscript{*}} 
      & -\phantom{\textsuperscript{*}} 
      & -\phantom{\textsuperscript{*}} 
      & -\phantom{\textsuperscript{*}} \\
    \quad Chain-of-Table~\citep{wang2024chainoftableevolvingtablesreasoning} 
      & PaLM 2 
      & 67.31\phantom{\textsuperscript{*}} 
      & 86.61\phantom{\textsuperscript{*}} 
      & -\phantom{\textsuperscript{*}} 
      & -\phantom{\textsuperscript{*}} 
      & -\phantom{\textsuperscript{*}} \\
    \quad Norm-DP\&Agent~\citep{liu2023rethinkingtabulardataunderstanding} 
      & GPT-3.5 
      & 73.65\phantom{\textsuperscript{*}} 
      & 88.50\phantom{\textsuperscript{*}} 
      & -\phantom{\textsuperscript{*}} 
      & -\phantom{\textsuperscript{*}} 
      & -\phantom{\textsuperscript{*}} \\
    \quad TIDE DP\&Agent~\citep{yang2025triples} 
      & GPT-3.5 
      & 75.00\phantom{\textsuperscript{*}} 
      & 89.82\phantom{\textsuperscript{*}} 
      & -\phantom{\textsuperscript{*}} 
      & -\phantom{\textsuperscript{*}} 
      & -\phantom{\textsuperscript{*}} \\
    \quad TableMaster~\citep{cao2025tablemasterrecipeadvancetable} 
      & GPT-4o-mini 
      & 78.13\phantom{\textsuperscript{*}} 
      & 90.12\phantom{\textsuperscript{*}} 
      & -\phantom{\textsuperscript{*}} 
      & 66.40\phantom{\textsuperscript{*}} 
      & -\phantom{\textsuperscript{*}} \\
    \quad E5~\citep{zhang-etal-2024-e5} 
      & GPT-4 
      & -\phantom{\textsuperscript{*}} 
      & -\phantom{\textsuperscript{*}} 
      & 85.08\phantom{\textsuperscript{*}} 
      & -\phantom{\textsuperscript{*}} 
      & -\phantom{\textsuperscript{*}} \\
    \quad SS-CoT~\citep{zhao2024stepwiseselfconsistentmathematicalreasoning} 
      & Llama-3.1$_{\text{70B}}$ 
      & 76.80\phantom{\textsuperscript{*}} 
      & -\phantom{\textsuperscript{*}} 
      & 79.10\phantom{\textsuperscript{*}} 
      & -\phantom{\textsuperscript{*}} 
      & -\phantom{\textsuperscript{*}} \\
    \midrule
    \multicolumn{7}{l}{\textit{\textbf{Finetuning-Based Methods}}} \\
    \quad FORTAP~\citep{cheng2022fortapusingformulasnumericalreasoningaware} 
      & BERT+LSTM 
      & -\phantom{\textsuperscript{*}} 
      & -\phantom{\textsuperscript{*}} 
      & 47.00\phantom{\textsuperscript{*}} 
      & -\phantom{\textsuperscript{*}} 
      & -\phantom{\textsuperscript{*}} \\
    \quad TAPEX-Large~\citep{liu2022tapextablepretraininglearning} 
      & BART$_{\text{Large}}$ 
      & 59.10\phantom{\textsuperscript{*}} 
      & 84.20\phantom{\textsuperscript{*}} 
      & 45.60\phantom{\textsuperscript{*}} 
      & -\phantom{\textsuperscript{*}} 
      & -\phantom{\textsuperscript{*}} \\
    \quad OmniTab~\citep{jiang-etal-2022-omnitab} 
      & BART$_{\text{Large}}$ 
      & 62.80\phantom{\textsuperscript{*}} 
      & -\phantom{\textsuperscript{*}} 
      & -\phantom{\textsuperscript{*}} 
      & -\phantom{\textsuperscript{*}} 
      & -\phantom{\textsuperscript{*}} \\
    \quad TableLlama~\citep{zhang2024tablellamaopenlargegeneralist} 
      & Llama-2$_{\text{7B}}$ 
      & 32.14\textsuperscript{*} 
      & 82.55\phantom{\textsuperscript{*}} 
      & 60.48\phantom{\textsuperscript{*}} 
      & 2.27\textsuperscript{*} 
      & 26.99\textsuperscript{*} \\
    \quad TableLLM~\citep{zhang2025tablellmenablingtabulardata} 
      & Qwen2$_{\text{7B}}$ 
      & 53.59\phantom{\textsuperscript{*}} 
      & 69.81\phantom{\textsuperscript{*}} 
      & 43.88\phantom{\textsuperscript{*}} 
      & 8.63\textsuperscript{*} 
      & 64.85\phantom{\textsuperscript{*}} \\
    \quad TableGPT2~\citep{su2024tablegpt2largemultimodalmodel} 
      & Qwen2.5$_{\text{7B}}$ 
      & 61.42\phantom{\textsuperscript{*}} 
      & 77.80\phantom{\textsuperscript{*}} 
      & 70.27\phantom{\textsuperscript{*}} 
      & 40.28\textsuperscript{*} 
      & 12.43\textsuperscript{*} \\
    \quad TabAF~\citep{wang2025generaltablequestionanswering}
      & Qwen2.5-Coder$_{\text{7B}}$
      & 74.72\phantom{\textsuperscript{*}}
      & 83.99\phantom{\textsuperscript{*}}
      & 78.41\phantom{\textsuperscript{*}}
      & 45.07\textsuperscript{*}
      & 62.33\textsuperscript{*} \\
    \rowcolor{gray!20}
    \quad \textbf{Formula-R1 (Ours)}
      & Qwen2.5-Coder$_{\text{7B}}$
      & \textcolor{langmildblue}{\textbf{75.27}}\phantom{\textsuperscript{*}}
      & \textcolor{langmildblue}{\textbf{85.08}}\phantom{\textsuperscript{*}}
      & \textcolor{langmildblue}{\textbf{78.84}}\phantom{\textsuperscript{*}}
      & \textcolor{langmildblue}{\textbf{70.10}}\phantom{\textsuperscript{*}}
      & \textcolor{langmildblue}{\textbf{83.50}}\textsuperscript{*} \\
    \rowcolor{gray!20}
    \quad \textbf{Formula-R1-Plus (Ours)}
      & Qwen2.5-Coder$_{\text{7B}}$
      & \textcolor{langdarkblue}{\textbf{87.53}}\phantom{\textsuperscript{*}}
      & \textcolor{langdarkblue}{\textbf{95.06}}\phantom{\textsuperscript{*}}
      & \textcolor{langdarkblue}{\textbf{89.83}}\phantom{\textsuperscript{*}}
      & \textcolor{langdarkblue}{\textbf{84.22}}\phantom{\textsuperscript{*}}
      & \textcolor{langdarkblue}{\textbf{95.53}}\textsuperscript{*} \\
    \bottomrule
    \end{tabular}%
  }
  \label{tab:comparison}
\end{table*}

\subsection{Performance of \textit{Formula-R1} and \textit{Formula-R1-Plus} Compared to Other Methods}

Table~\ref{tab:comparison} presents the performance of \textit{Formula-R1} and \textit{Formula-R1-Plus} compared to several strong baselines, as detailed in Appendix~\ref{ap:detail_setting}. \textit{Formula-R1} is derived from the best overall performance achieved with \textit{Formula Tuning} under the cold-start RL setting.

For a fair experimental comparison, we adopt the self-consistency strategy \citep{wang2023selfconsistencyimproveschainthought} to enhance table understanding performance following prior work \citep{liu2023rethinkingtabulardataunderstanding, yang2025triples, wang2025generaltablequestionanswering}. This strategy generates multiple candidate formulas and selects the final answer through majority voting, and has been widely adopted as an effective approach for improving accuracy in table reasoning. Specifically \textit{Formula-R1} follows previous work~\citep{wang2025generaltablequestionanswering} by generating 10 symbolic reasoning outputs. To further leverage the complementary strengths of textual and symbolic reasoning, we introduce \textit{Formula-R1-Plus}, which produces a balanced set of 5 textual and 5 symbolic outputs. \textit{Formula-R1-Plus} incorporates both textual and symbolic reasoning components trained with RL, where the textual component is optimized with RL from a cold start using binary answer correctness as the reward signal.
\\

\noindent\textbf{RL with cold start SFT improves numerical reasoning over SFT alone.}  
TabAF~\citep{wang2025generaltablequestionanswering} is a strong baseline that employs SFT for symbolic reasoning via spreadsheet formulas and similarly adopts a hybrid self-consistency strategy with 5 textual and 5 formula-based outputs. Nevertheless, \textit{Formula-R1-Plus} significantly outperforms TabAF, demonstrating that RL offers clear advantages over SFT-only models, even when those models are distilled from stronger teacher models. These gains are particularly pronounced on benchmarks that require substantial numerical reasoning and complex table understanding, such as binary verification in \textit{TabFact}, multi-step numerical computation in \textit{FinQA}, and in-the-wild complex table reasoning in \textit{AIT-QA}, where \textit{Formula-R1-Plus} surpasses the SFT-based TabAF by approximately 12, 35, and 29 points in accuracy, respectively.
\\

\noindent\textbf{\textit{Formula-R1-Plus} delivers consistently strong performance.}
\textit{Formula-R1-Plus} surpasses all finetuning-based methods across the reported datasets. Specifically, it achieves 80.47\% on FinQA, demonstrating strong complex mathematical reasoning ability. On AIT-QA, \textit{Formula-R1-Plus} brings an improvement of 30.9 points, highlighting the out-of-distribution robustness enabled by RL. These results also show that smaller open-source models can outperform larger closed-source models. Despite using a 7B-parameter \textit{Qwen} backbone, \textit{Formula-R1-Plus} achieves competitive performance with—and outperforms several—prompting-based methods, including GPT-4o-mini-based TableMaster and GPT-4-based E5. The original \textit{Formula-R1}, which relies solely on formula-based reasoning, also achieves competitive performance across benchmarks.

Besides, \textit{Formula-R1} achieves 40.85\% on MultiHiertt and 35.22\% on TableBench, while \textit{Formula-R1-Plus} achieves 51.73\% and 44.96\%, respectively.  
\\

\begin{table*}[t!]
  \centering
  \caption{Performance under Zero-shot, SFT, and RL settings across models and datasets. Values in the table indicate accuracy (\%). \textit{Text} and \textit{Formula} refer to textual and symbolic reasoning methods, respectively.  The best performance in each column is highlighted in \textcolor{langdarkblue}{dark blue}, and the second-best in \textcolor{langmildblue}{light blue}.}
  % \textit{w/ CS} denotes RL initialized from cold-start SFT.
  \resizebox{1\textwidth}{!}{%
    \begin{tabular}{llcccccccc}
    \toprule
    \multirow[c]{2}{*}{\textbf{Base Model}}  
      & \multirow[c]{2}{*}{\textbf{Method}}  
      & \multicolumn{5}{c}{\textbf{In‐Distribution}}  
      & \multicolumn{2}{c}{\textbf{Out‐of‐Distribution}}
      & \multirow[c]{2}{*}{\textbf{Overall}} \\
    \cmidrule(lr){3-7}\cmidrule(lr){8-9}
    & & \textbf{WikiTQ} & \textbf{TabFact} & \textbf{FinQA} & \textbf{HiTab} & \textbf{MultiHiertt}  
      & \textbf{AIT-QA} & \textbf{TableBench} & \\
    \midrule
    \multicolumn{10}{l}{\textit{\textbf{Zero‐Shot}}} \\
    \noalign{\vskip 2pt}
    \multirow[c]{2}{*}{\quad Llama-3.1$_{\text{8B}}$}   
      & Text      & 50.46 & 67.84 & 42.37 & 29.92 & 18.77  
                  & 59.61 & 21.29 & 41.47 \\
      & Formula   & 12.58 & 15.77 & 22.67 &  6.57 &  4.23  
                  & 20.00 & 10.31 & 13.16 \\
    \cmidrule(lr){2-10}
    \multirow[c]{2}{*}{\quad Qwen2.5-Coder$_{\text{7B}}$}
      & Text      & 55.53 & 78.26 & 55.54 & 50.38 & 26.11  
                  & 73.60 & 24.12 & 51.93 \\
      & Formula   & 38.96 & 52.52 & 34.44 & 32.60 & 15.90  
                  & 52.43 & 26.05 & 36.13 \\
    \midrule
    \multicolumn{10}{l}{\textit{\textbf{Supervised Fine‐Tuning (SFT)}}} \\
    \multirow[c]{2}{*}{\quad Llama-3.1$_{\text{8B}}$}   
      & Text      & 66.15 & 82.95 & 50.04 & 70.27 & 39.98  
                  & 78.45 & 23.10 & 58.71 \\
      & Formula   & 59.62 & 72.48 & 58.50 & 72.54 & 44.00  
                  & 74.95 & 31.48 & 59.08 \\
    \cmidrule(lr){2-10}
    \multirow[c]{2}{*}{\quad Qwen2.5-Coder$_{\text{7B}}$}
      & Text      & 65.98 & 81.13 & 59.46 & 72.35 & 43.93  
                  & 81.55 & 24.01 & 61.20 \\
      & Formula   & 63.46 & 78.16 & 58.33 & 71.83 & 42.68  
                  & 76.12 & 35.56 & 60.88 \\
    \midrule
    
    \multicolumn{10}{l}{\textit{\textbf{Reinforcement Learning (RL)}}} \\
    \multirow[c]{2}{*}{\quad Llama-3.1$_{\text{8B}}$}   
      & Text          
          & 64.37 & 82.16 & 62.60 & 68.81 & 31.99  
          & 82.91 & 28.08 & 60.13 \\
      & Formula       
          & 57.64 & 80.09 & 60.85 & 67.93 & 29.40  
          & 80.78 & 30.69 & 58.20 \\
    \cmidrule(lr){2-10}
    \multirow[c]{2}{*}{\quad Qwen2.5-Coder$_{\text{7B}}$}
      & Text          
          & 66.95 & 85.43 & 64.34 & 74.24 & 35.55  
          & 85.28 & 27.86 & 62.80 \\
      & Formula       
          & 67.80 & 84.19 & 62.16 & 71.19 & 41.72  
          & 81.17 & 35.45 & 63.38 \\
    \midrule
    
     \multicolumn{10}{l}{\textit{\textbf{RL with Cold-Start SFT (RL w/ CS)}}} \\
     \multirow[c]{2}{*}{\quad Llama-3.1$_{\text{8B}}$}   
      & Text
          & \textcolor{langdarkblue}{\textbf{71.56}} & \textcolor{langdarkblue}{\textbf{87.01}} & 56.84 & 77.64 & 49.34  
          & \textcolor{langdarkblue}{\textbf{85.66}} & 28.16 & 65.17 \\
      % \rowcolor{gray!20}
      & Formula
          & 70.49 & 83.04 & \textcolor{langdarkblue}{\textbf{71.99}} & \textcolor{langdarkblue}{\textbf{79.29}} & \textcolor{langmildblue}{\textbf{54.55}}  
          & 81.29 & \textcolor{langmildblue}{\textbf{36.64}} & \textcolor{langmildblue}{\textbf{68.18}} \\
    \cmidrule(lr){2-10}
     \multirow[c]{2}{*}{\quad Qwen2.5-Coder$_{\text{7B}}$}
      & Text    
          & \textcolor{langmildblue}{\textbf{71.31}} & 86.07 & 64.77 & 77.42 & 54.25  
          & \textcolor{langmildblue}{\textbf{85.43}} & 25.94 & 66.46 \\
      % \rowcolor{gray!20}
      & Formula
          & 70.90 & \textcolor{langmildblue}{\textbf{86.18}} & \textcolor{langmildblue}{\textbf{69.21}} & \textcolor{langmildblue}{\textbf{77.89}} & \textcolor{langdarkblue}{\textbf{56.78}}  
          & 79.14 & \textcolor{langdarkblue}{\textbf{39.25}} & \textcolor{langdarkblue}{\textbf{68.48}} \\
    \bottomrule
    \end{tabular}%
  }
  \label{tab:main_perf}
  % \vspace{-10pt}
\end{table*}

\subsection{Performance of Formula Learning under SFT and RL}

Table~\ref{tab:main_perf} presents the performance of different formula learning methods under supervised fine-tuning (\textit{SFT}), reinforcement learning (\textit{RL}), reinforcement learning with a cold-start SFT initialization (\textit{RL w/ CS}), and direct zero-shot inference without any task-specific training. This experiment is primarily designed to compare RL and SFT, as well as textual and symbolic reasoning, in table understanding scenarios. Several key findings and insights are summarized below:
\\

\noindent\textbf{A large zero-shot gap exists between textual and symbolic reasoning.}  
Under zero-shot settings, both \textit{Llama-3.1${}_{8\text{B}}$} and \textit{Qwen2.5-Coder${}_{7\text{B}}$} show a substantial gap between textual and formula-based reasoning. While textual reasoning achieves moderate performance, symbolic reasoning via spreadsheet formulas remains much weaker, especially for \textit{Llama-3.1${}_{8\text{B}}$}. Although \textit{Qwen2.5-Coder${}_{7\text{B}}$}, likely benefiting from code pretraining, exhibits relatively stronger zero-shot formula generation ability, the overall results still suggest that vanilla pretraining does not adequately equip small open-source LMs with spreadsheet syntax and symbolic reasoning skills.
\\

\noindent\textbf{SFT effectively injects formula knowledge and rapidly closes the text-formula gap}  
After SFT, formula-based reasoning improves dramatically for both backbones, narrowing the gap with textual reasoning to a very small margin. For example, on \textit{Llama-3.1${}_{8\text{B}}$}, the overall performance of formula reasoning rises from 13.16\% in zero-shot to 59.08\% after SFT, nearly matching textual reasoning at 58.71\%. A similar trend is observed for \textit{Qwen2.5-Coder${}_{7\text{B}}$}, where formula reasoning reaches 60.88\%, close to the 61.20\% achieved by textual reasoning. These results show that SFT serves as an effective knowledge injection stage, enabling small models to acquire basic spreadsheet formula generation and symbolic reasoning ability.
\\

\noindent\textbf{RL further improves performance, with stronger gains for symbolic reasoning and out-of-distribution generalization.}  
Compared with SFT, RL brings further improvements, especially for formula-based reasoning. For \textit{Qwen2.5-Coder${}_{7\text{B}}$}, formula reasoning improves from 60.88\% under SFT to 63.38\% under RL, surpassing textual reasoning at 62.80\%. Moreover, the gains on out-of-distribution benchmarks such as \textit{AIT-QA} and \textit{TableBench} are particularly notable, suggesting that RL improves not only answer correctness but also the model's robustness under distribution shifts. In contrast, textual reasoning benefits less consistently from RL, indicating that execution-based reinforcement is especially well suited to symbolic formula generation.
\\

\noindent\textbf{Cold-start SFT initialization is crucial for maximizing RL performance.}  
Initializing RL from an SFT-trained model consistently yields the strongest results across both backbones. Compared with RL from scratch, the \textit{RL w/ CS} setting provides substantial additional gains for both textual and symbolic reasoning. In particular, the formula-based variant under \textit{RL w/ CS} achieves the best overall performance among all open-source settings, reaching 68.18\% on \textit{Llama-3.1${}_{8\text{B}}$} and 68.48\% on \textit{Qwen2.5-Coder${}_{7\text{B}}$}. These findings indicate that SFT provides a strong initialization by injecting symbolic prior knowledge, while RL further unlocks the model's reasoning potential through execution-based optimization.
\\

\noindent\textbf{Symbolic reasoning demonstrates clear advantages on numerically intensive and structurally complex benchmarks.}  
The benefits of formula-based reasoning become particularly evident on benchmarks that require numerical computation or complex table manipulation. Under \textit{RL w/ CS}, formula reasoning achieves the best results on \textit{FinQA}, \textit{HiTab}, \textit{MultiHiertt}, and \textit{TableBench} across both backbones. For example, on \textit{Qwen2.5-Coder${}_{7\text{B}}$}, formula reasoning reaches 69.21\% on \textit{FinQA}, 77.89\% on \textit{HiTab}, 56.78\% on \textit{MultiHiertt}, and 39.25\% on \textit{TableBench}, consistently outperforming the corresponding textual reasoning results. These results confirm that explicit symbolic reasoning is particularly advantageous for tasks involving multi-step numerical computation, hierarchical structure understanding, and complex table operations.
\\

\noindent\textbf{Textual reasoning remains competitive on simpler verification or direct lookup-style tasks.}  
Despite the strong overall performance of symbolic reasoning, textual reasoning remains preferable on some relatively simpler tasks. In particular, under \textit{RL w/ CS}, textual reasoning achieves the best scores on \textit{WikiTQ}, \textit{TabFact}, and \textit{AIT-QA} for at least one backbone. For example, \textit{Llama-3.1${}_{8\text{B}}$} with textual reasoning attains 87.01\% on \textit{TabFact} and 85.66\% on \textit{AIT-QA}, outperforming its formula-based counterpart. This suggests that when answers can be directly inferred, verified, or extracted without complex symbolic composition, textual reasoning may remain more efficient and effective.
\\

\subsection{Comparative Analysis of Spreadsheet Formulas, SQL, and Python for Symbolic Table Reasoning}
\label{ap:other_symbolic}

% \begin{figure}[t]
%   \centering
%   \begin{subfigure}[t]{0.4\textwidth}
%     \centering
%     \includegraphics[height=120pt]{figures/symbolic_zs.pdf}
%     \caption{Zero-Shot setting}
%     \label{fig:symbolic_zs}
%   \end{subfigure}
%   \begin{subfigure}[t]{0.4\textwidth}
%     \centering
%     \includegraphics[height=120pt]{figures/symbolic_rl.pdf}
%     \caption{Reinforcement Learning setting}
%     \label{fig:symbolic_rl}
%   \end{subfigure}
%   \caption{Comparison of performance across different symbolic reasoning methods under Zero-Shot and Reinforcement Learning (RL) settings. Values in the figure indicate accuracy (\%).}
%   \label{fig:symbolic_comparison}
%    % Big thanks to blingbing for the figure of double radar plots!
% \end{figure}

In this section, we analyze the use of formulas as a symbolic reasoning tool, and compare them with SQL and Python.  
% Figure~\ref{fig:symbolic_comparison} and 
Table~\ref{tab:symbolic} compare three symbolic reasoning paradigms (SQL queries, Python snippets, and spreadsheet formulas) under both zero-shot and reinforcement learning (RL) settings. To ensure a fair comparison, we train and evaluate these symbolic tools only on datasets with flat, relational tables: WikiTQ, TabFact, and FinQA for training, and all three plus TableBench for evaluation. This restriction is necessary because SQL operates solely on flat tables, while Python/Pandas are also usually worked for flat tabular structures. Several patterns emerge from the comparison:

\textbf{Spreadsheet formulas offer the strongest zero-shot symbolic reasoning.}  
Without any task-specific training, formulas achieve the highest out-of-the-box performance, with an overall accuracy of 37.99\%. They slightly outperform Python (35.64\%) and significantly surpass SQL (11.84\%). The gap is especially pronounced on datasets like WikiTQ and FinQA, suggesting that pre-trained language models possess some intuitive understanding of spreadsheet-style operations, but struggle to generate valid and executable SQL or Python code without further adaptation. On TableBench, which features complex table QA questions, Python code falls short. Models without fine-tuning often fail to generate long and sufficiently accurate code to solve challenging problems. In contrast, spreadsheet formulas are shorter, easier to generate, and more robust in zero-shot settings, making them better suited for this type of reasoning.

\begin{wraptable}{r}{0.5\textwidth} 
% \begin{table}[ht]
  \centering
  \caption{Performance comparison of different symbolic reasoning methods (SQL, Python, and Formula) under Zero-shot and RL settings. Values in the table indicate accuracy (\%). The best performance in each column is highlighted in \textcolor{langdarkblue}{dark blue}.}
  \resizebox{0.5\textwidth}{!}{%
    \begin{tabular}{lccccc}
    \toprule
    \textbf{Method} & \textbf{WikiTQ} & \textbf{TabFact} & \textbf{FinQA} & \textbf{TableBench} & \textbf{Overall} \\
    \midrule
    \multicolumn{6}{l}{\textit{\textbf{Zero-Shot}}} \\
    \quad SQL     & 17.07 & 21.15 & 1.51  & 7.64  & 11.84 \\
    \quad Python  & 30.96 & \textcolor{langdarkblue}{\textbf{60.39}} & \textcolor{langdarkblue}{\textbf{34.87}} & 16.33 & 35.64 \\
    \quad \textbf{Formula}
      & \textcolor{langdarkblue}{\textbf{38.96}} 
      & 52.52 
      & 34.44
      & \textcolor{langdarkblue}{\textbf{26.05}} 
      & \textcolor{langdarkblue}{\textbf{37.99}} \\
    \midrule
    \multicolumn{6}{l}{\textit{\textbf{Reinforcement Learning (RL)}}} \\
    \quad SQL     & 67.58 & 83.94 & 38.79 & 32.09 & 55.60 \\
    \quad Python  & 70.46 & 84.42 & 65.85 & 35.26 & 64.00 \\
    \quad \textbf{Formula}
      & \textcolor{langdarkblue}{\textbf{70.67}} 
      & \textcolor{langdarkblue}{\textbf{84.50}} 
      & \textcolor{langdarkblue}{\textbf{65.89}} 
      & \textcolor{langdarkblue}{\textbf{35.84}} 
      & \textcolor{langdarkblue}{\textbf{64.23}} \\
    \bottomrule
    \end{tabular}%
  }
  \label{tab:symbolic}
% \end{table}
\end{wraptable}

\textbf{Spreadsheet formulas remain the most effective symbolic tool after RL.}  
All three symbolic tools improve significantly with RL training, with SQL and Python gaining 43.8 and 28.4 percentage points, respectively. This underscores the value of policy-gradient optimization for learning execution-constrained program structures. Post-training, formulas and Python reach nearly identical accuracy (64.23\% vs.\ 64.00\%), while SQL still lags behind at 55.60\%, largely due to its limited ability to handle numerical computation required in table reasoning. Although the final scores of formula-based and Python-based reasoning are close, formula-based reasoning achieves performance comparable to Python, with a slight relative advantage. Beyond their strong performance, spreadsheet formulas are shorter, easier to read, and more beginner-friendly. These qualities make them not only effective but also practical and accessible as a symbolic tool for table reasoning tasks.

\subsection{Additional results and further analysis.}  

A \textbf{reward sensitivity analysis} is presented in Appendix~\ref{ap:reward_sensitivity}, where we further examine the choice of 0.2 as the partial reward value in our main experiments.
An \textbf{ablation study} and upper-bound performance analysis of \textit{Formula Tuning} are presented in Appendix~\ref{ap:ablation}.  
An impact analysis of the \textbf{reasoning process} during formula tuning appears in Appendix~\ref{ap:reasoning_process}.
A \textbf{statistical analysis} of the generated formulas is provided in Appendix~\ref{ap:stat}, and qualitative \textbf{case studies} are discussed in Appendix~\ref{ap:case_study}.

\section{Conclusion}
In this paper, we introduced \textit{Formula Tuning} (Fortune), an RL framework that trains LLMs to generate executable spreadsheet formulas for table understanding tasks. Based on this framework, we developed \textit{Formula-R1} and demonstrated that formula-driven RL can substantially enhance the symbolic reasoning capabilities of LLMs over complex tables with numerical computation. Extensive experiments across multiple benchmarks show that our approach is particularly effective for numerical computation and complex table reasoning, while also yielding strong out-of-distribution generalization. Our findings highlight the promise of formula-driven learning as a practical and effective paradigm for improving LLM reasoning on tabular tasks.

\clearpage

% \subsubsection*{Broader Impact Statement}
% In this optional section, TMLR encourages authors to discuss possible repercussions of their work,
% notably any potential negative impact that a user of this research should be aware of. 
% Authors should consult the TMLR Ethics Guidelines available on the TMLR website
% for guidance on how to approach this subject.

% in many journals. This is optional and at the discretion of the authors. Only add
% this information once your submission is accepted and deanonymized. 

% Only add this information once your submission is accepted and deanonymized. 

\bibliography{main}
\bibliographystyle{tmlr}

%%%%%%%%%%%%%%%%%%%%%%%%%%%%%%%%%%%%%%%%%%%%%%%%%%%%%%%%%%%%

\clearpage
\DoToC
\clearpage

\appendix

\section{Limitations}
Although Fortune achieves promising improvements in table reasoning, several limitations remain. First, the current training data could be further expanded with more complex and diverse examples to improve generalization. Second, due to limited computational resources, our experiments are based on single training runs without variance estimates, leaving the statistical significance of the observed improvements uncertain. These limitations suggest promising future directions, including scaling Formula Tuning to broader domains, incorporating related formula-based tasks, improving the robustness of reinforcement learning, and developing better reward designs for structured output generation.

\section{Ethical Statement}
\label{ap:ethic}

% \noindent\textbf{Broader Impacts.}  
\textit{Formula Tuning} (Fortune) introduces a reinforcement learning framework that enhances symbolic reasoning for table understanding via spreadsheet formulas. By improving the ability of language models to reason over tabular data with verifiable, executable outputs, our work offers substantial benefits in domains where transparency and precision are essential—such as education, scientific analysis, finance, and public policy. Executable formulas can provide interpretable and auditable reasoning steps, potentially increasing user trust and reliability in AI-generated decisions involving structured data.

However, these capabilities also introduce potential risks. If applied carelessly, formula generation may amplify biases present in training data or propagate subtle numerical errors. Moreover, spreadsheet formulas are deeply embedded in productivity workflows, and inaccurate generation at scale could lead to downstream harms (e.g., miscalculated budgets or flawed data reports). Furthermore, since symbolic reasoning via formulas may be more accessible in high-resource languages or domains with well-structured spreadsheets, deployment in low-resource settings could exacerbate inequalities in model performance and accessibility.

% \noindent\textbf{Safeguards.}  
To mitigate such risks, we recommend several safeguards for future use of Fortune and similar symbolic reasoning systems. First, generated formulas should undergo verification through deterministic execution engines to ensure correctness. Second, evaluations should be conducted across diverse domains and spreadsheet structures, particularly including noisy or adversarial formats. Third, human-in-the-loop validation should be used in high-stakes applications (e.g., healthcare or financial audits) to ensure interpretability and safety. Finally, we advocate for transparent reporting of formula generation limitations and the inclusion of provenance indicators that show how a particular output was derived, enabling error tracing and accountability.

\section{Supplementary Discussion of Methodology}
\label{ap:discussion}

\subsection{Textual vs. Symbolic Reasoning in Table Understanding}

% In addition to the formal analysis in Section~\ref{sec:text_symbolic}, we highlight several conceptual advantages of symbolic reasoning for table understanding:
In this section, we highlight several conceptual advantages of symbolic reasoning for table understanding:

\begin{itemize}[leftmargin=1.5em, itemsep=0.5em]
    \item \textbf{Execution-based computation.} Symbolic reasoning externalizes computation through deterministic execution, separating high-level logical planning from low-level arithmetic or formatting operations.

    \item \textbf{Compositionality and structure.} Spreadsheet formulas offer compositional and type-aware representations, providing stronger structural priors than unstructured text.

    \item \textbf{Verifiability and transparency.} Symbolic outputs are interpretable and verifiable: they can be inspected, tested, reused, or debugged—enabling traceable and auditable reasoning processes.

    \item \textbf{Discrete action space.} The symbolic action space is bounded and discrete, which facilitates more stable exploration and optimization during training.

    \item \textbf{Robustness to token-level variability.} Unlike textual reasoning, which is prone to errors from exposure bias or numerical drift, symbolic reasoning delegates exact computation to the executor, reducing dependency on fragile token generation.
\end{itemize}

\subsection{SFT vs. RL in Symbolic Table Reasoning}

We also expand upon the discussion, comparing supervised fine-tuning (SFT) and reinforcement learning (RL) for symbolic reasoning:
% We also expand upon the discussion in Section~\ref{sec:sft_rl}, comparing supervised fine-tuning (SFT) and reinforcement learning (RL) for symbolic reasoning:

\begin{itemize}[leftmargin=1.5em, itemsep=0.5em]
    \item \textbf{SFT limitations.} SFT imitates teacher demonstrations at the token level and struggles to generalize beyond the training distribution. It penalizes semantically correct but structurally different formulas, constraining exploration.

    \item \textbf{Reward-aligned optimization.} RL optimizes directly for task-level correctness using execution-based rewards, allowing the model to discover diverse yet valid solution strategies.

    \item \textbf{Support for many-to-one mappings.} Since different formulas can yield the same correct answer, RL naturally accommodates this multiplicity, whereas SFT often fails to reward such diversity.

    \item \textbf{Flexible reward shaping.} RL allows for auxiliary reward terms—such as penalties on length, syntactic constraints, or correctness under verification—which are difficult to incorporate in SFT.

    \item \textbf{Improved generalization.} By optimizing for semantic correctness rather than mimicking surface-level token patterns, RL enables the model to generalize more effectively in both in-distribution (ID) and out-of-distribution (OOD) scenarios, including novel question types, unseen table schemas, and structurally diverse formulas.
\end{itemize}

\subsection{Practical Challenges of Formula Tuning}
\label{ap:challenge}

While reinforcement learning (RL) offers significant advantages for symbolic table reasoning, it also introduces several practical challenges.

% , especially under the \textbf{assumptions} outlined in Section~\ref{sec:text_symbolic} and \ref{sec:sft_rl}.

\begin{itemize}[leftmargin=1.5em, itemsep=0.3em]
    \item \textbf{Exploration bottlenecks.} We assumes that the RL policy can eventually explore correct formulas. However, the space of possible formulas is extremely large, and valid, executable ones are rare—especially at the start of training. This makes it difficult for the model to receive useful reward signals, leading to slow or unstable learning.
    % \item \textbf{Exploration bottlenecks.} Assumption~\ref{as:rl_exploration} assumes that the RL policy can eventually explore correct formulas. However, the space of possible formulas is extremely large, and valid, executable ones are rare—especially at the start of training. This makes it difficult for the model to receive useful reward signals, leading to slow or unstable learning.

    \item \textbf{Limited symbolic priors.} Unlike supervised fine-tuning (SFT), RL does not benefit from direct examples of correct formulas. If the model lacks prior knowledge of spreadsheet syntax or symbolic structures, it may struggle to generate meaningful outputs. This weak starting point often results in inefficient exploration and poor early performance.

    \item \textbf{RL training instability.} When training from scratch, the model often produces repetitive, invalid, or meaningless formulas in the early stages, receiving no reward. This can cause unstable training and hinder convergence. Empirically, initializing with a supervised or pretrained model leads to more stable training and faster reward learning.

    \item \textbf{Sparse and coarse reward signals.} Execution-based rewards typically only indicate whether the final answer is correct or not, without offering any feedback on partially correct or structurally promising outputs. This makes it harder for the model to learn from near misses. Designing more informative reward functions—such as those based on formula structure or partial execution—remains an important direction.
\end{itemize}

Overcoming these challenges is essential for scaling \textit{Formula Tuning} to more complex symbolic tasks, broader domains, and higher-capacity models. Future work may explore techniques such as curriculum learning, hybrid supervision, symbolic inductive priors, or multi-objective optimization to improve training stability and exploration efficiency.

\section{Detailed Settings of Experiments}
\label{ap:detail_setting}

\noindent\textbf{Models.}
Our experiments include both open-source and proprietary models. For open-source models, we use \textit{Qwen2.5-Coder$_{7B}$} (\texttt{Qwen2.5-Coder-7B-Instruct}, Apache 2.0 License) \citep{hui2024qwen25codertechnicalreport} and \textit{LLaMA-3.1$_{8B}$} (\texttt{LLaMA-3.1-8B-Instruct}, Meta Llama 3 Community Licence) \citep{grattafiori2024llama3herdmodels}.  
% For proprietary models, we evaluate OpenAI’s\footnote{\url{https://openai.com/policies/row-terms-of-use/}} \textit{GPT-4o} (\texttt{gpt-4o-2024-11-20}), \textit{GPT-4o-mini} (\texttt{gpt-4o-mini-2024-07-18}), and \textit{O1} (\texttt{o1-2024-12-17}) as baselines.

\noindent\textbf{Datasets.}
As shown in Table~\ref{tab:datasets}, we conduct experiments on seven diverse table understanding benchmarks: WikiTQ~\citep{pasupat2015compositionalsemanticparsingsemistructured}, TabFact~\citep{chen2020tabfactlargescaledatasettablebased}, FinQA~\citep{chen-etal-2021-finqa}, HiTab~\citep{cheng2021hitab}, MultiHiertt~\citep{zhao-etal-2022-multihiertt}, AIT-QA~\citep{katsis2021aitqa}, and TableBench~\citep{wu2025tablebenchcomprehensivecomplexbenchmark}. These datasets vary in domain coverage, table structures, and question complexity, collectively spanning the full spectrum of table understanding tasks. For MultiHiertt, which contains multiple tables, we concatenate them vertically to form a single spreadsheet-like table. For training, we combine the first five datasets into a unified training corpus and train the model jointly on this merged set. Each dataset is then evaluated individually. All original training and test splits are preserved, except for TabFact, from which we randomly sample 10,000 examples to prevent its abundance of relatively simple binary QA examples from dominating or skewing the training. Among these benchmarks, AIT-QA and TableBench are considered out-of-distribution (OOD) evaluation sets, while the remaining datasets are treated as in-distribution (ID). The characteristics of each dataset are summarized below:

\begin{itemize}[leftmargin=20pt, itemsep=0pt, labelsep=5pt, topsep=0pt]
  \item \textbf{WikiTQ} is a Wikipedia-based table QA dataset with relatively simple factoid questions over relational tables.
  \item \textbf{TabFact} also uses Wikipedia tables but frames the task as fact verification, where each claim is labeled as either \textit{true} or \textit{false}.
  \item \textbf{FinQA} focuses on financial-domain tables and requires symbolic reasoning over semi-structured input that includes both pretext and posttext as additional context.
  \item \textbf{HiTab} contains hierarchical tables derived from statistical reports. While its structure is more complex than relational tables, the content is relatively straightforward.
  \item \textbf{MultiHiertt} involves multi-table reasoning over hierarchical tables in the financial domain, demanding both structural and symbolic reasoning.
  \item \textbf{AIT-QA} consists of hierarchical tables from the airline domain. Although structurally rich, its questions tend to be simpler.
  \item \textbf{TableBench} features complex questions over relational tables drawn from various domains. Many questions require multi-step symbolic reasoning, making it the most challenging benchmark in our evaluation.
\end{itemize}

We merge the training sets of WikiTQ, TabFact, FinQA, HiTab, and MultiHiertt into a corpus of 45,198 examples for RL training. After GPT-4o distillation and rejection filtering to remove demonstrations with incorrect answers, 31,840 examples remain for formula-based SFT. The cold-start SFT stage uses 10,000 examples drawn from the five training datasets before subsequent RL training.

\begin{table*}[ht]
\centering
\caption{Overview of the training data and table benchmarks used in this study.}
\resizebox{\textwidth}{!}{%
\begin{tabular}{llccccccl}
\toprule
Evaluation Type        & Dataset & \# Train Data & \# Test Data & Table Type                   & Domain              & License & Source \\
\midrule
\multirow{5}{*}{In-Distribution}
  & WikiTQ~\citep{pasupat2015compositionalsemanticparsingsemistructured}
    & 13{,}753 & 4{,}217 & Relational             & Wikipedia           & CC-BY-SA-4.0 & \href{https://github.com/ppasupat/WikiTableQuestions}{Link} \\
  & TabFact~\citep{chen2020tabfactlargescaledatasettablebased}
    & 10{,}000 & 2{,}024 & Relational             & Wikipedia           & CC-BY-4.0    & \href{https://github.com/wenhuchen/Table-Fact-Checking}{Link} \\
  & FinQA~\citep{chen-etal-2021-finqa}
    & 6{,}251  & 1{,}147 & Relational             & Finance             & MIT          & \href{https://github.com/czyssrs/FinQA}{Link} \\
  & HiTab~\citep{cheng2021hitab}
    & 7{,}399  & 1{,}583 & Hierarchical           & Statistical Reports & C-UDA 1.0    & \href{https://github.com/microsoft/HiTab}{Link} \\
  & MultiHiertt~\citep{zhao-etal-2022-multihiertt}
    & 7{,}795  & 1{,}038 & Multiple \& Hierarchical & Finance             & MIT          & \href{https://github.com/psunlpgroup/MultiHiertt}{Link} \\
\midrule
\multirow{2}{*}{Out-of-Distribution}
  & AIT-QA~\citep{katsis2021aitqa}
    & --    & 515  & Hierarchical           & Airline             & CDLA-Sharing-1.0 & \href{https://github.com/IBM/AITQA}{Link} \\
  & TableBench~\citep{wu2025tablebenchcomprehensivecomplexbenchmark}
    & --    & 883  & Relational             & Cross Domain        & CC0-1.0       & \href{https://huggingface.co/datasets/Multilingual-Multimodal-NLP/TableBench}{Link} \\
\bottomrule
\end{tabular}
} % end resizebox
\label{tab:datasets}
\end{table*}

\noindent\textbf{Table Encoding.}  
We adopt a table encoding method similar to SpreadsheetEncoder~\citep{dong-etal-2024-encoding}, which converts a table into a linearized markdown-style format. Each cell is represented by its spreadsheet address and value, forming text sequences such as \texttt{A1,Year|A2,Profit}. This encoding preserves both structural and content information, enabling the model to better understand cell-level references.

\noindent\textbf{Output Format.}  
Following the structured reasoning paradigm, the model is required to produce outputs in a two-stage format:

\[
  y = 
  \underbrace{\langle\texttt{think}\rangle\, t \,\langle/\texttt{think}\rangle}_{\textit{\textbf{reasoning trajectory}}}\,
  \underbrace{\langle\texttt{answer}\rangle\, \texttt{\{json\}} \,\langle/\texttt{answer}\rangle}_{\textit{\textbf{final answer}}},
\]

where \(t\) is a free-form natural language reasoning process (i.e., the thinking process), and the answer block contains a JSON object from which the final prediction is extracted. This design enables decoupling the reasoning trajectory from the answer payload and facilitates more structured reward computation.

To encourage adherence to this format, we introduce a lightweight \textit{format reward}. If the output fails to follow the required structure (e.g., malformed tags or unparseable JSON), the model receives a penalty of \(-2\). If the format is valid and the answer can be successfully parsed from the JSON object, a small positive reward of \(+0.1\) is added to the answer-level reward. Therefore, the final reward is as:
\begin{equation}
r_{\text{final}}(a \mid s) = r_{\text{ans}}(a \mid s) + r_{\text{fmt}}(a)
\end{equation}
This reward shaping helps stabilize training and guide the model toward producing reliably structured outputs.

\noindent\textbf{Baselines.}
We compare our proposed framework against a broad range of strong baselines, including both prompting-based and fine-tuning-based methods. To ensure a fair comparison, we require all methods to output short, deterministic answers rather than open-ended free-form text. Following this criterion, we exclude TableLLM~\citep{zhang2025tablellmenablingtabulardata}, which relies on a critique model for answer evaluation and does not produce a directly verifiable answer string. Prompting-based methods currently dominate the TableQA landscape, with most relying on large closed-source models for performance. We compare \textit{Formula-R1} and \textit{Formula-R1-Plus} with several representative methods in this category: Binder~\citep{cheng2023bindinglanguagemodelssymbolic}, Dater~\citep{ye2023largelanguagemodelsversatile}, API-Assisted~\citep{cao2025tablemasterrecipeadvancetable}, Chain-of-Table~\citep{wang2024chainoftableevolvingtablesreasoning}, ReAcTable~\citep{zhang2023reactableenhancingreacttable}, Norm-DP~\citep{liu2023rethinkingtabulardataunderstanding}, TIDE~\citep{yang2025triples}, E5~\citep{zhang-etal-2024-e5}, and SS-CoT~\citep{zhao2024stepwiseselfconsistentmathematicalreasoning}. We also include TableMaster~\citep{cao2025tablemasterrecipeadvancetable}, a recent recipe-based prompting framework built on GPT-4o-mini. For fine-tuning-based methods, we select models specifically trained for table question answering tasks. These include TAPEX-Large~\citep{liu2022tapextablepretraininglearning}, OmniTab~\citep{jiang-etal-2022-omnitab}, TableLlama~\citep{zhang2024tablellamaopenlargegeneralist}, TableGPT2~\citep{su2024tablegpt2largemultimodalmodel}, and TabAF~\citep{wang2025generaltablequestionanswering}, a recent strong method that combines formula generation and hybrid self-consistency. Our framework is evaluated under the same settings to ensure consistency and comparability across methods.

\noindent\textbf{RL Training.}
We use Proximal Policy Optimization (PPO) for reinforcement learning. The maximum prompt and response lengths are set to 8,192 and 512 tokens, respectively. The critic is initialized with the same weights as the actor. We use learning rates of $1 \times 10^{-6}$ and $1 \times 10^{-5}$ for the actor and critic, respectively. The KL-divergence coefficient is set to 0.001 to balance exploration and policy stability. During training, we use a sampling temperature of 0.6 to promote diversity in the generated reasoning chains and formula outputs. The PPO minibatch size is 64. We hold out 2.5\% of the training data as a development set for checkpoint selection.

\paragraph{Reward formulation.}
For formula-based reasoning, the final scalar reward is:
\begin{equation}
r_{\mathrm{final}}(y,s)=
\begin{cases}
-2, & \text{if the output format is invalid}, \\
0, & \text{if the format is valid but formula execution fails}, \\
0.3, & \text{if execution succeeds but the answer is incorrect}, \\
1.1, & \text{if execution succeeds and the answer is correct}.
\end{cases}
\end{equation}
Here, a valid format requires the prescribed reasoning and answer tags
and a parseable JSON object containing the formula. For executable
formulas, the answer reward is $1$ for a correct answer and $0.2$ for
an incorrect answer, with an additional $0.1$ format reward.
If execution fails, the final reward is $0$, without the format bonus.
Invalid-format outputs receive $-2$, and their answer reward is not evaluated.

\noindent\textbf{SFT Training.}
We construct the supervised fine-tuning (SFT) corpus by prompting \textit{GPT-4o} with ground-truth answers to generate chain-of-thought reasoning followed by a final answer. We adopt a rejection-based fine-tuning (RFT) strategy~\citep{yuan2023scalingrelationshiplearningmathematical}, retaining only examples for which the generated answer exactly matches the ground truth. For symbolic reasoning tasks, correctness is determined by executing the generated formula and verifying that its output matches the expected answer. This procedure provides high-quality supervision for fine-tuning. All SFT models are trained for up to four epochs with a learning rate of $2 \times 10^{-5}$ and a batch size of 64. We hold out 2.5\% of the training data as a development set for checkpoint selection.

% , and we report results based on the checkpoint with the highest exact match accuracy on the test set

\noindent\textbf{Evaluation Inference.}  
For all models, including both open-source and proprietary ones, we use a temperature of 0, top-$k$ of 50, and top-$p$ of 0.7 during inference. Setting the temperature to 0 encourages deterministic outputs and improves stability in single-pass predictions. For self-consistency decoding in \textit{Formula-R1}, we use a higher temperature of 0.6 to promote diversity across multiple samples, enabling the model to better explore reasoning variations and improve final answer voting.

\noindent\textbf{Evaluation Metrics.}  
Following prior work~\citep{pasupat2015compositionalsemanticparsingsemistructured, cheng2021hitab}, we primarily use exact match (EM) as the evaluation metric, applying numeric tolerance when comparing numerical values. Official evaluation scripts are used whenever available to ensure consistency. For TabFact, which is formulated as a binary classification task, we report standard classification accuracy. Since our training objective aligns with evaluation, we also use exact match (EM) for answer reward calculation. For voting, we follow the answer-level frequency voting principle of TabAF~\citep{wang2025generaltablequestionanswering}: we execute generated formulas to obtain candidate answers, combine them with directly generated textual answers when applicable, group equivalent answers, and select the most frequent answer.

\noindent\textbf{Software.}  
We implement Fortune using Python 3.11, with the VERL framework~\citep{sheng2024hybridflow} serving as the core architecture for reinforcement learning and other supervised fine-tuning with language models. Our implementation utilizes VLLM (v0.8.3)~\citep{kwon2023efficientmemorymanagementlarge} for efficient LLM inference and generation, PyTorch (v2.4.0) with CUDA 12.8 for deep learning operations, and Ray~\citep{moritz2018raydistributedframeworkemerging} for distributed training and inference. FlashAttention-2~\citep{dao2022flashattentionfastmemoryefficientexact} is integrated to accelerate attention computation. For proprietary LLMs, we access OpenAI models via the Microsoft Azure platform\footnote{\url{https://azure.microsoft.com/}}. For formula execution, we use the open-source spreadsheet engine \texttt{formulas}\footnote{\url{https://github.com/vinci1it2000/formulas}} (EUPL 1.1+ License), which supports a wide range of standard spreadsheet operators. A representative list of symbolic operators used in our table reasoning framework is provided in Appendix~\ref{ap:operator}.

\noindent\textbf{Hardware.}  
All experiments are conducted on machines equipped with either NVIDIA A100 80GB PCIe or NVIDIA H100 80GB PCIe GPUs, along with 1.0 TB of RAM. For reinforcement learning (RL) training of open-source models, we use 8 × NVIDIA H100 80GB PCIe GPUs by default. For supervised fine-tuning (SFT) of open-source models, we use 4 × NVIDIA A100 80GB PCIe GPUs by default.

\section{Cold-Start Data Scaling}
\label{sec:cold_start_scaling}

The cold-start supervised fine-tuning (SFT) stage is used to provide the model with basic formula-based reasoning capabilities before reinforcement learning (RL). Based on our empirical experience, approximately 500 high-quality SFT examples are generally sufficient for the model to acquire the basic reasoning schema required by our framework, including spreadsheet syntax, cell referencing, arithmetic operations, aggregation functions, and other fundamental formula-construction rules. This initialization substantially facilitates subsequent RL training by improving early exploration and reducing the frequency of invalid or meaningless formula generations.

However, increasing the amount of SFT data does not necessarily lead to better RL performance. While additional SFT examples may further improve the initial policy, excessive teacher supervision can over-constrain the policy toward the teacher distribution, reduce the diversity of generated formulas, and limit exploration of alternative valid solutions. This may in turn reduce the performance ceiling that can be achieved through RL. In contrast, RL can explore multiple executable formulas that produce the same correct answer, which is particularly important because the formula action space is many-to-one.

Therefore, the role of cold-start SFT is to provide sufficient initialization for stable exploration rather than to maximize imitation performance. In our setting, approximately 500 high-quality examples serve as an empirical reference point for acquiring the basic formula-based reasoning schema. This value should not be interpreted as a universal minimum threshold, since the required amount may depend on the backbone model, the diversity and quality of the distilled data, and the complexity of the target tasks. A more systematic study of the trade-off between cold-start coverage, policy diversity, and RL performance is an important direction for future work.

\section{Reward Sensitivity Analysis}
\label{ap:reward_sensitivity}

In this section, we further analyze the sensitivity of \textit{Formula Tuning} to the reward design. In particular, we focus on the partial reward assigned to \emph{executable but incorrect} formulas. This intermediate reward is introduced to alleviate reward sparsity. In our preliminary experiments, using only binary rewards $\{1, 0\}$ led to less stable optimization and substantially lower performance. We therefore introduce an intermediate reward for executable but incorrect formulas to provide denser learning feedback. A key reason is that small pretrained language models typically lack sufficient prior knowledge of spreadsheet-formula syntax, making it difficult to reliably optimize answer correctness before the model first learns to produce valid and executable formulas.

To study the effect of this design choice, we vary the partial reward assigned to executable-but-incorrect outputs while keeping all other settings fixed. Specifically, we conduct RL-only training with \textit{Qwen2.5-Coder$_{7\text{B}}$} on HiTab. The results are reported in Table~\ref{tab:reward_sensitivity}.

\begin{table}[h!]
\centering
\small
\caption{Sensitivity analysis of the partial reward assigned to executable-but-incorrect outputs on HiTab using \textit{Qwen2.5-Coder$_{7\text{B}}$} under RL-only training.}
\label{tab:reward_sensitivity}
\begin{tabular}{lccccc}
\toprule
Partial reward (exec-but-wrong) & 0.0 & 0.1 & 0.2 & 0.3 & 0.4 \\
\midrule
Accuracy & 57.98 & 70.64 & 72.32 & 72.04 & 70.53 \\
\bottomrule
\end{tabular}
\end{table}

Several observations can be drawn from these results. First, the model is highly sensitive to removing partial credit entirely: when the reward for executable-but-incorrect outputs is set to 0.0, performance drops substantially to 57.98. This confirms that a purely binary reward is insufficient for stable optimization in symbolic table reasoning. Second, performance remains strong and relatively stable when the partial reward lies within a moderate range of 0.1 to 0.3, with the best result achieved at 0.2. This indicates that the method is not overly sensitive to the exact value of the partial reward, as long as a reasonable amount of intermediate credit is provided. Rather than biasing the model toward incorrect outputs, the partial reward serves as a meaningful intermediate learning signal that encourages exploration of valid formulas. Third, assigning an excessively large partial reward (e.g., 0.4) leads to a slight performance drop, suggesting that overly rewarding executability without sufficient emphasis on correctness may weaken the model's incentive to optimize toward the final answer.

Overall, these findings support the reward design in \textit{Formula Tuning}. The intermediate reward for executable-but-incorrect formulas improves optimization stability and helps the model acquire symbolic structure before answer correctness is fully optimized. At the same time, the results suggest that the method is fairly robust to the precise choice of this reward within a reasonable range. This design is particularly important in our setting, where valid formula generation itself is a nontrivial capability for small language models.

\section{Formula--Python Comparison on Hierarchical Tables}

\begin{table}[h]
  \centering
  \caption{Zero-shot accuracy (\%) on hierarchical and multiple-table benchmarks. Overall is the arithmetic mean across the three benchmarks. The best performance in each column is highlighted in \textcolor{langdarkblue}{dark blue}.}
  \resizebox{0.4\textwidth}{!}{%
    \begin{tabular}{lcccc}
      \toprule
      \textbf{Method} & \textbf{HiTab} & \textbf{MultiHiertt} & \textbf{AIT-QA} & \textbf{Overall} \\
      \midrule
      Python & 28.46 & 8.36 & 45.22 & 27.35 \\
      \textbf{Formula} & \textcolor{langdarkblue}{\textbf{32.60}} & \textcolor{langdarkblue}{\textbf{15.90}} & \textcolor{langdarkblue}{\textbf{52.43}} & \textcolor{langdarkblue}{\textbf{33.64}} \\
      \bottomrule
    \end{tabular}%
  }
  \label{tab:symbolic_hierarchical}
\end{table}

We extend the zero-shot comparison to HiTab, MultiHiertt, and AIT-QA,
which contain hierarchical or multiple-table structures.
As shown in Table~\ref{tab:symbolic_hierarchical}, Formula achieves
higher accuracy than Python on all three benchmarks, extending
the evidence beyond our previous flat-table comparison.

\section{Ablation Study and Upper-Bound Performance of \textit{Formula-R1}}
\label{ap:ablation}
\begin{table*}[ht!]
  \centering
  \caption{Performance comparison of TabAF, \textit{Formula-R1}, and \textit{Formula-R1-Plus} variants. Values indicate accuracy (\%). `-' indicates results not reported in the related paper. Upper-bound results are shown in gray. The best non-oracle performance in each column is highlighted in \textcolor{langdarkblue}{dark blue}.}
  \resizebox{\textwidth}{!}{%
    \setlength{\aboverulesep}{1pt}
    \setlength{\belowrulesep}{1pt}
    \begin{tabular}{llccccccc}
      \toprule
      \multicolumn{2}{l}{\textbf{Method}}
        & \textbf{WikiTQ} & \textbf{TabFact} & \textbf{FinQA}
        & \textbf{HiTab} & \textbf{MultiHiertt}
        & \textbf{AIT-QA} & \textbf{TableBench} \\
      \midrule
      \multirow{4}{*}{TabAF \citep{wang2025generaltablequestionanswering}}
        & \textcolor{gray}{Upper Bound}
        & \textcolor{gray}{80.13} & \textcolor{gray}{94.02} & \textcolor{gray}{-}
        & \textcolor{gray}{82.07} & \textcolor{gray}{-}
        & \textcolor{gray}{-} & \textcolor{gray}{-} \\
        & 5 Text
        & 61.42 & 81.47 & - & 74.24 & - & - & - \\
        & 5 Formula
        & 64.20 & 67.54 & - & 74.87 & - & - & - \\
        & 5 Text + 5 Formula
        & 74.72 & 83.99 & - & 78.41 & - & - & - \\
      \midrule
      \multirow{2}{*}{\textbf{Formula-R1}}
        & \textcolor{gray}{Upper Bound}
        & \textcolor{gray}{77.35} & \textcolor{gray}{96.49} & \textcolor{gray}{72.01}
        & \textcolor{gray}{79.91} & \textcolor{gray}{58.86}
        & \textcolor{gray}{88.93} & \textcolor{gray}{43.26} \\
        & 10 Formula
        & 75.27 & 85.08 & 70.10 & 78.84 & 58.48 & 83.50 & 42.13 \\
      \midrule
      \multirow{2}{*}{\textbf{Formula-R1-Plus}}
        & \textcolor{gray}{Upper Bound}
        & \textcolor{gray}{93.62} & \textcolor{gray}{99.51} & \textcolor{gray}{91.02}
        & \textcolor{gray}{95.01} & \textcolor{gray}{71.29}
        & \textcolor{gray}{98.06} & \textcolor{gray}{61.16} \\
        & 5 Text + 5 Formula
        & \textcolor{langdarkblue}{\textbf{87.53}}
        & \textcolor{langdarkblue}{\textbf{95.06}}
        & \textcolor{langdarkblue}{\textbf{84.22}}
        & \textcolor{langdarkblue}{\textbf{89.83}}
        & \textcolor{langdarkblue}{\textbf{68.40}}
        & \textcolor{langdarkblue}{\textbf{95.53}}
        & \textcolor{langdarkblue}{\textbf{47.23}} \\
      \bottomrule
    \end{tabular}%
  }
  \label{tab:ablation}
\end{table*}

We conduct an ablation study of \textit{Formula-R1-Plus} to investigate the complementary roles and effectiveness of textual and symbolic reasoning. Table~\ref{tab:ablation} reveals several notable patterns.

\textbf{Combining textual and symbolic reasoning yields the most robust performance.}  
The balanced sampling strategy adopted by \textit{Formula-R1-Plus} (five textual and five formula-based candidates) consistently outperforms both the pure-text and pure-formula variants across all benchmarks. This result suggests that textual and symbolic reasoning address complementary error modes. Disabling either modality leads to substantial accuracy drops, with degradations of up to 18 percentage points (e.g., 18 points on FinQA for text-only and 13 points on AIT-QA for formula-only).

\textbf{Textual and formula-based reasoning excel in different scenarios.}  
Textual reasoning performs better on relatively simpler table QA tasks, such as TabFact and AIT-QA, where natural language understanding and logical inference play a dominant role. In contrast, formula-based reasoning is more effective on arithmetic-intensive or structurally complex tasks such as FinQA and MultiHiertt, where symbolic execution is essential for deriving the correct answer. This contrast further highlights the importance of integrating both modalities for general-purpose table understanding.

\textbf{Reinforcement learning enhances symbolic reasoning beyond supervised fine-tuning.}  
Compared with TabAF, which uses the same backbone but is trained solely with SFT, the RL-based variant of \textit{Formula-R1} achieves substantially stronger formula-only performance (e.g., 82.41\% vs.\ 67.54\% on TabFact with 5 Formula). This result suggests that reinforcement learning encourages the model to explore more reliable and executable reasoning paths, ultimately improving the quality of symbolic programs.

\textbf{A substantial number of correct answers are lost under naive majority voting.}  
The \textit{upper-bound} rows show that \textit{Formula-R1-Plus} frequently generates correct answers that are not selected by simple majority voting. The gap between the upper bound and actual performance reaches 19 points on MultiHiertt and 17 points on TableBench, indicating considerable headroom for improved candidate selection through confidence-based aggregation or more effective reranking strategies.

\textbf{Structurally complex and low-resource benchmarks remain the most challenging.}  
The largest performance gaps appear on datasets such as MultiHiertt and TableBench, which are both structurally complex and relatively low-resource. These results highlight the limitations of current voting and reasoning mechanisms, and point to promising future directions such as symbolic planner integration, adaptive sampling, and confidence-calibrated answer selection.

\section{Impact Analysis of the Thinking Process in Formula Tuning}
\label{ap:reasoning_process}

\begin{figure*}[t!]
    \centering
    \includegraphics[width=\textwidth]{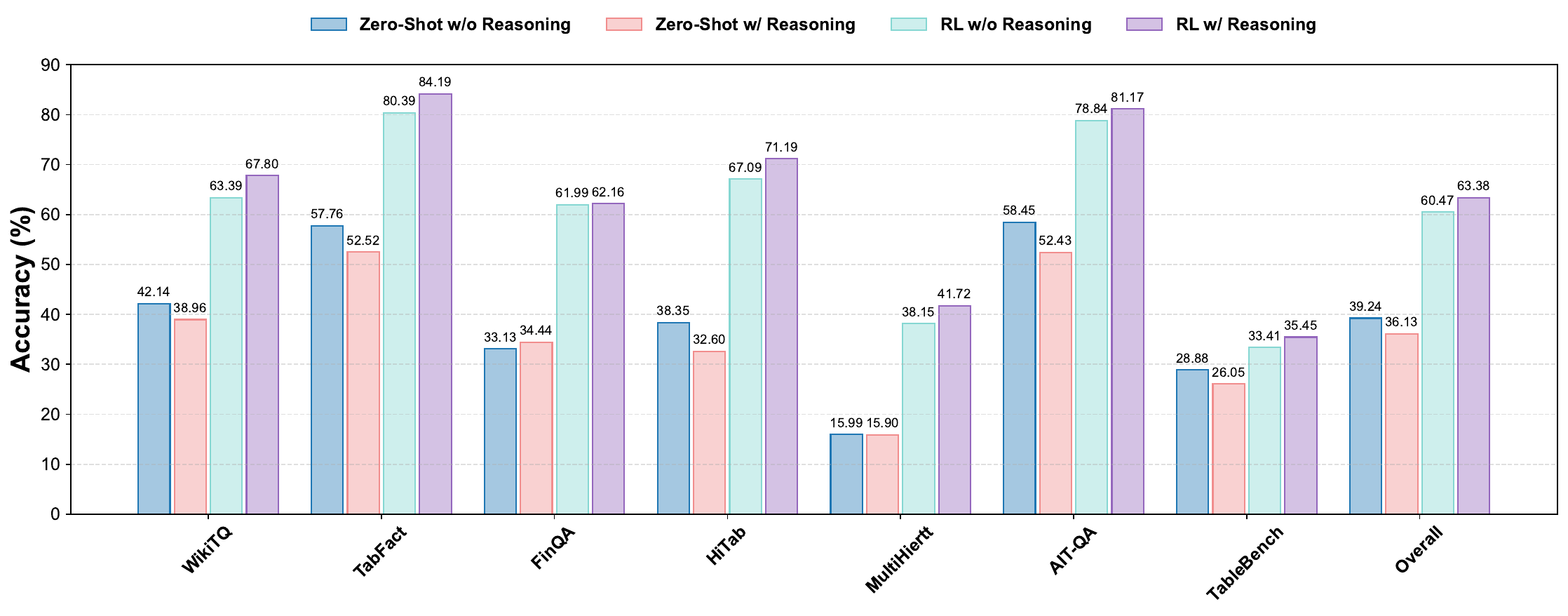}
    \caption{Performance comparison with and without explicit reasoning process under Zero-Shot and Reinforcement Learning (RL) settings across various datasets. Each group of bars shows the accuracy (\%) achieved by four configurations: Zero-Shot without Reasoning, Zero-Shot with Reasoning, RL without Reasoning, and RL with Reasoning.}
    \label{fig:reasoning_impact}
    % Acknowledging Jingyu for the figure A.
\end{figure*}

Table~\ref{tab:wo_reasining} and Figure~\ref{fig:reasoning_impact} highlight the impact of incorporating an explicit thinking process before generating formulas. 

\begin{table*}[ht!]
  \centering
  \caption{Performance comparison with and without reasoning under Zero-Shot and RL settings across various datasets. Values in the table indicate accuracy (\%).}
    \resizebox{0.9\textwidth}{!}{%
    \begin{tabular}{lccccccc c}
    \toprule
    \textbf{Method} & \textbf{WikiTQ} & \textbf{TabFact} & \textbf{FinQA} & \textbf{HiTab} & \textbf{MultiHiertt} & \textbf{AIT-QA} & \textbf{TableBench} & \textbf{Overall} \\
    \midrule
    \multicolumn{9}{l}{\textit{\textbf{Zero Shot}}} \\
    \quad w/o Reasoning & 42.14 & 57.76 & 33.13 & 38.35 & 15.99 & 58.45 & 28.88 & 39.24 \\
    \quad w/ Reasoning  & 38.96 & 52.52 & 34.44 & 32.60 & 15.90 & 52.43 & 26.05 & 36.13 \\
    \midrule
    \multicolumn{9}{l}{\textit{\textbf{Reinforcement Learning (RL)}}} \\
    \quad w/o Reasoning & 63.39 & 80.39 & 61.99 & 67.09 & 38.15 & 78.84 & 33.41 & 60.47 \\
    \quad w/ Reasoning  & 67.80 & 84.19 & 62.16 & 71.19 & 41.72 & 81.17 & 35.45 & 63.38 \\
    \bottomrule
    \end{tabular}%
  }
  \label{tab:wo_reasining}
\end{table*}

\textbf{In the zero-shot setting, reasoning may hurts performance.}  
We observe that adding a reasoning trace before the formula generally leads to lower accuracy (e.g., 39.24\% → 36.13\% overall). This is likely because such \textit{thought-first-then-formula} generation patterns are underrepresented in pretraining corpora. As a result, models tend to produce unnatural or error-prone reasoning steps, which negatively affect the final output. The limitations of chain-of-thought reasoning in certain scenarios have also been discussed in recent work~\citep{sprague2025cotcotchainofthoughthelps,liu2024mindstepbystep}.

\textbf{In the RL setting, reasoning significantly improves performance.}  
Once trained with our answer-based reward, the model begins to benefit from generating an explicit reasoning trace. The inclusion of a thinking process effectively expands the exploration space during policy optimization and encourages the model to break down complex table reasoning tasks into more manageable steps. This leads to consistent performance improvements across datasets (e.g., 60.47\% → 63.38\% overall), demonstrating that reasoning becomes a valuable asset—once the model has been properly trained to utilize it effectively.

\begin{figure*}[t!]
    \centering
    \includegraphics[width=0.85\textwidth]{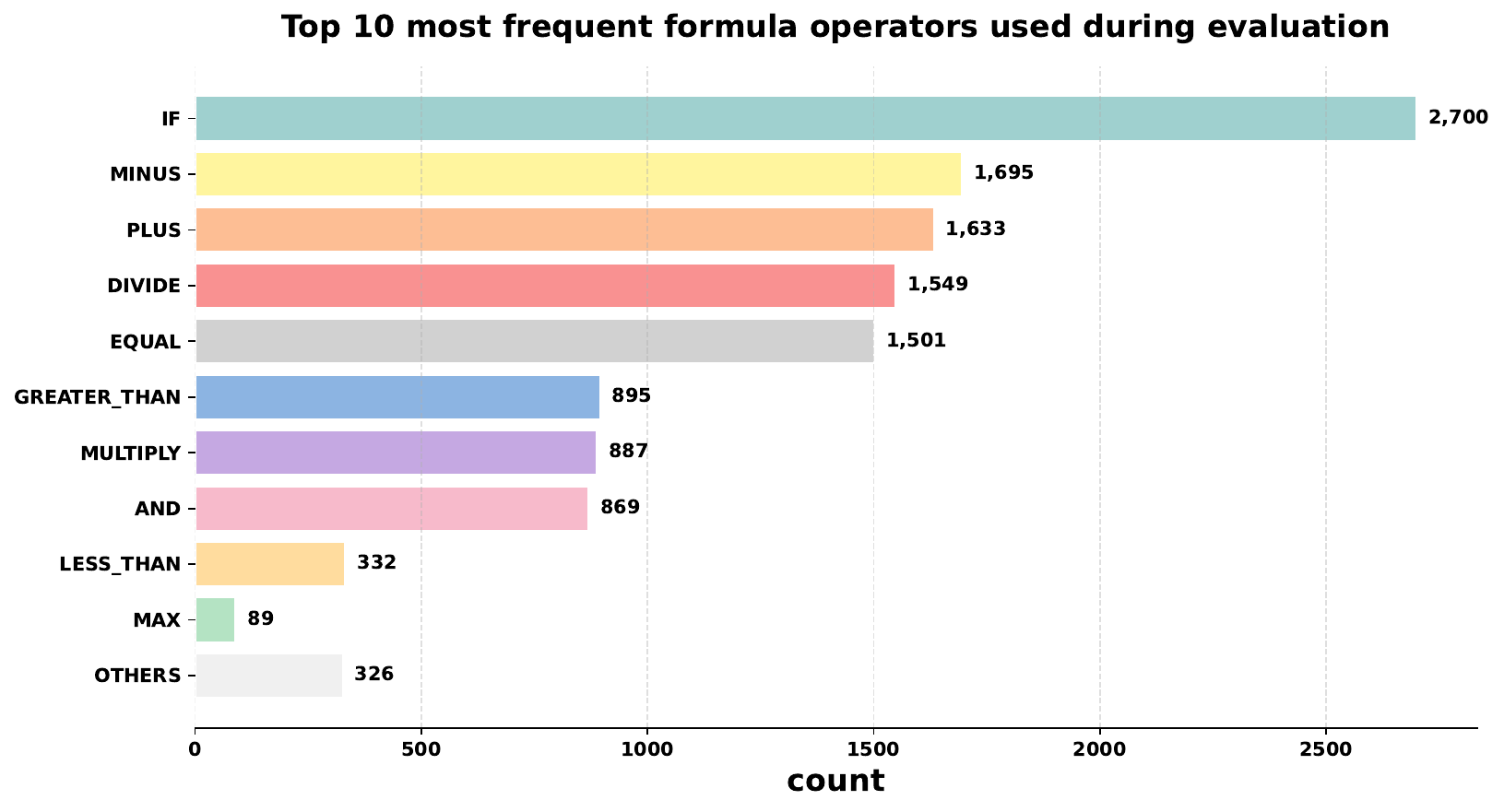}
    \caption{Top 10 most frequent formula operators used during evaluation.}
    \label{fig:top10_formula}
    % Acknowledging Jingyu for the figure B.
\end{figure*}

\section{Statistical Analysis of Generated Formulas in Symbolic Table Reasoning}
\label{ap:stat}

We collect and analyze statistics of the formulas generated by our formula-tuned model during evaluation to better understand their structural properties.

\begin{table*}[ht]
\caption{Statistics of generated formulas across different table understanding datasets.}
\centering
\resizebox{0.7\textwidth}{!}{%
\begin{tabular}{lcccccc}
\toprule
\textbf{Dataset} 
& \multicolumn{3}{c}{\textbf{Table Layout}} 
& \multicolumn{3}{c}{\textbf{Generated Formula}} \\
\cmidrule(lr){2-4} \cmidrule(lr){5-7}
& \textbf{Width} & \textbf{Height} & \textbf{Area Size} 
& \textbf{Length} & \textbf{\# Operators} & \textbf{\# Variables} \\
\midrule
WikiTQ       & 6.28 & 19.46 & 121.68 & 21.61 & 0.84 & 1.50 \\
TabFact      & 6.28 & 14.04 & 87.42  & 44.85 & 3.11 & 2.85 \\
FinQA        & 3.92 & 18.09 & 70.80  & 18.84 & 1.98 & 2.94 \\
HiTab        & 6.16 & 21.32 & 171.40 & 12.20 & 0.49 & 1.34 \\
MultiHiertt  & 7.25 & 46.59 & 339.30 & 22.64 & 2.10 & 3.21 \\
AIT-QA       & 5.62 & 13.86 & 81.91  & 5.67  & 0.11 & 1.89 \\
TableBench   & 6.71 & 16.26 & 108.24 & 26.87 & 1.52 & 2.78 \\
\bottomrule
\end{tabular}
}
\label{tab:formula_stats}
\end{table*}

Table~\ref{tab:formula_stats} presents a quantitative summary of both table layout characteristics and the structural properties of generated formulas across seven widely used table understanding datasets. The table is divided into two parts: the first group (\textit{Width}, \textit{Height}, and \textit{Area Size}) reflects the average structural complexity of the input tables, while the second group (\textit{Length}, \textit{\# Operators}, and \textit{\# Variables}) captures the syntactic and symbolic complexity of the generated formulas.

We observe substantial variation in table layout complexity. For example, MultiHiertt has by far the largest average table area (339.30), indicating its multiple and hierarchical format. In contrast, datasets like AIT-QA and FinQA involve relatively smaller or simpler tables, which may place less structural burden on the reasoning process. Notably, HiTab also exhibits a high area size, despite having fewer variables and a short formula length, suggesting that its challenge lies more in table structure than in formula richness.

In terms of generated formulas, TabFact stands out with the longest average formula length (44.85 characters) and the highest number of operators (3.11), indicating that its fact verification tasks typically require complex symbolic conditions. On the other hand, AIT-QA exhibits the shortest formulas with minimal operator usage (5.67 length, 0.11 operators), reflecting the dataset’s relatively simple question types. Datasets like FinQA and MultiHiertt show high variable counts (around 3 per formula), which aligns with their multi-step reasoning nature involving multiple cell references. TableBench poses a greater challenge due to its combination of complex question intent and compositional reasoning demands. Although its average table size is moderate, the questions often require multi-step symbolic operations such as nested aggregations, comparisons, or indirect references—making it a strong testbed for evaluating deep reasoning ability.

These statistics provide important insights for model evaluation and reward design. First, different datasets pose very different reasoning demands—relying solely on benchmarks like WikiTQ or AIT-QA may underestimate a model’s true symbolic capacity. Second, symbolic complexity (e.g., operator density) varies nontrivially across tasks, and therefore reward shaping mechanisms should adapt accordingly to avoid penalizing semantically necessary long formulas. Lastly, the disconnect between table area and formula length in datasets like HiTab implies that structural layout, rather than size alone, can be the main source of reasoning difficulty—an insight that can guide future benchmark construction and curriculum learning design.

Figure~\ref{fig:top10_formula} presents the distribution of the most frequently used formula operators during evaluation. The conditional operator \texttt{IF} appears overwhelmingly often, with a count of 2,700, indicating that conditional reasoning is central to many table reasoning tasks. Arithmetic operators such as \texttt{MINUS} (1,695), \texttt{PLUS} (1,633), and \texttt{DIVIDE} (1,549) are also widely used, reflecting the numerical nature of many questions. Logical comparison operators like \texttt{EQUAL}, \texttt{GREATER\_THAN}, and \texttt{LESS\_THAN} occur frequently as well, suggesting that relational reasoning is also a common requirement. Less frequently used operators such as \texttt{MAX} and those grouped into the \texttt{OTHERS} category play a smaller role. Overall, the operator distribution highlights the need for models to support both arithmetic and logical reasoning, with a strong emphasis on conditional operations.

\clearpage
\section{Case Study}
\label{ap:case_study}

\subsection{Textual vs. Symbolic Reasoning}
We present representative examples comparing textual and symbolic reasoning methods in table understanding tasks.

As shown in Table~\ref{tab:example_text}, the textual approach performs better in this particular case. This is a simple counting question, so textual reasoning can easily enumerate the relevant items and output the correct answer (4). In contrast, the symbolic reasoning attempts to solve the problem via a more complex formula. Although the reasoning process is logically correct and the intent aligns with expectations, the actual formula execution produces an incorrect result due to implementation details—specifically, the presence of the string \textit{nan} in the table being misinterpreted. This highlights that in certain corner cases, symbolic reasoning may not have fully mastered tool usage or aligned formula execution. In comparison, textual reasoning can sometimes bypass such pitfalls and arrive at the correct answer more robustly.

On the other hand, Table~\ref{tab:example_symbolic} illustrates a case where symbolic reasoning proves more effective. This is a moderately difficult classic counting problem with a ground truth answer of 13. Here, textual reasoning fails, producing an incorrect count (22), suggesting that its performance deteriorates as task complexity increases. In contrast, symbolic reasoning correctly formulates a count-based expression and arrives at the correct answer. The accompanying reasoning trace (\textit{Think}) also clearly articulates the logic behind the computation, demonstrating the advantage of symbolic approaches in precise numerical tasks.

\begin{table*}[h]
\centering
\caption{Case study comparing textual and symbolic reasoning. In this example, textual reasoning yields the correct answer.}
\small
\begin{tabular}{p{0.15\textwidth}p{0.8\textwidth}}
\toprule
\textbf{Method} & \textbf{Case} \\
\midrule
- & \textbf{Question:} \textit{rapper saigon has had at least () albums released in the U.S.?} \\
& \textbf{Table:} \\
& 
\includegraphics[width=0.81\textwidth]{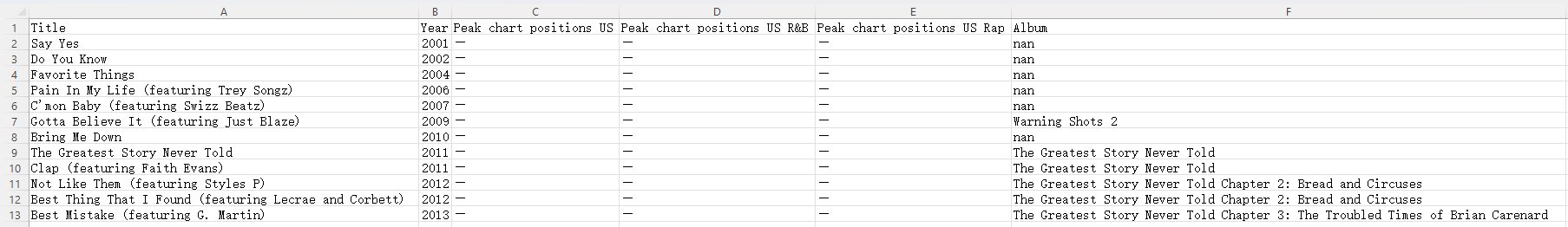}
\\
& \textbf{Groundtruth Answer:} 4 \\

\midrule
Symbolic & \textbf{Think:} \textit{To compute how many unique albums Saigon has released, I examine column F (F2:F13), which lists album names. Some of these cells contain "nan", which I treat as missing values. I only count the unique non-missing values. A standard formula to count unique non-empty values is: =SUMPRODUCT(1/COUNTIF(F2:F13,F2:F13)) This formula works because COUNTIF returns the number of times each value appears; taking 1 over that and summing gives the count of distinct values. This will count duplicates only once and ignore empty cells or "nan" if treated as the same placeholder.} \\
& \textbf{Generated Formula:} \texttt{=SUMPRODUCT(1/COUNTIF(F2:F13,F2:F13))} \\
\\
& \textbf{Execution Result:} 5 \\
\\
& \textbf{Execution Accuracy:} 0.0 \\

\midrule
Textual & \textbf{Think:} \textit{We are asked to find out how many albums rapper Saigon has released in the U.S. From the table, we look at the 'Album' column (column F), which lists the associated album for each song. We ignore entries that are 'nan' or empty, as those do not correspond to a known album. From F2 to F13, we extract the non-nan values: - Warning Shots 2 - The Greatest Story Never Told - The Greatest Story Never Told Chapter 2: Bread and Circuses - The Greatest Story Never Told Chapter 3: The Troubled Times of Brian Carenard These are 4 distinct album names, so the answer is 4.} \\
& \textbf{Generated Answer:} \texttt{4} \\
\\
& \textbf{Answer Accuracy:} 1.0 \\
\bottomrule
\end{tabular}
\label{tab:example_text}
\end{table*}

\begin{table*}[ht]
\centering
\caption{Case study comparing textual and symbolic reasoning. In this example, symbolic reasoning yields the correct answer.}
\small
\begin{tabular}{p{0.15\textwidth}p{0.8\textwidth}}
\toprule
\textbf{Method} & \textbf{Case} \\
\midrule
- & \textbf{Question:} \textit{How many elections did the party win75 or more seats?} \\
& \textbf{Table:} \\
& 
\includegraphics[width=0.35\textwidth]{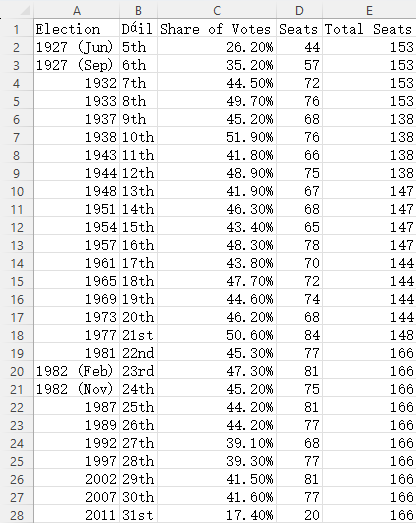}
\\
& \textbf{Groundtruth Answer:} 13\\

\midrule
Textual & \textbf{Think:} \textit{- A2, 1927 (sep): 57 seats - A3, 1932: 72 seats - ... - A26, 2002: 81 seats - A27, 2007: 77 seats - A28, 2011: 20 seats (less than 75) Counting the rows where the "seats" column is 75 or more: - 1927 (sep): 57 - 1932: 72 - ... - 2002: 81 - 2007: 77 Counting these, we find 22 elections where the party won 75 or more seats.} \\
& \textbf{Generated Answer:} \texttt{22} \\
\\
& \textbf{Execution Accuracy:} 0.0 \\

\midrule
Symbolic & \textbf{Think:} \textit{1. To answer this question, I need to look into the table provided in the question. The question is asking to determine how many times a party won 75 or more seats in various elections based on a given table. 2. Since the question is focused on the number of seats won, I need to look for a column in the table that contains the number of seats each party won in each election. 3. Given that the question is asking to count the number of elections where the party won 75 or more seats, I will need to look for a column that contains the seat numbers and then count how many of those seat numbers are 75 or greater. 4. Since the question is not providing a specific column in the table for seat numbers, I will need to infer that the seat numbers are contained in column D of the provided table, which is titled "seats".
} \\
& \textbf{Generated Formula:} \texttt{=COUNTIF(D2:D28, ">=75")} \\
\\
& \textbf{Execution Result:} 13 \\
\\
& \textbf{Execution Accuracy:} 1.0 \\
\bottomrule
\end{tabular}
\label{tab:example_symbolic}
\end{table*}

\clearpage
\subsection{Performance of SFT vs. RL}
We also compare models trained with supervised fine-tuning (SFT) and reinforcement learning (RL) in the context of symbolic table reasoning using formulas. As shown in Table~\ref{tab:example_rl}, RL enables the language model to generate more accurate and executable formulas, leading to improved symbolic reasoning performance. In this example, the SFT model appears to merely memorize a reasoning pattern and produces a formula that is superficially related to the question but logically incorrect. Moreover, the intermediate reasoning process is highly interpretable and clearly aligned with the question's logic.

\begin{table*}[h]
\centering
\caption{Case study comparing performance of SFT vs. RL. In this example, RL yields the correct answer.}
\small
\begin{tabular}{p{0.15\textwidth}p{0.8\textwidth}}
\toprule
\textbf{Method} & \textbf{Case} \\
\midrule
- & \textbf{Question:} \textit{What is the number of winning drivers that represented the U.S.?} \\
& \textbf{Table:} \\
& 
\includegraphics[width=0.81\textwidth]{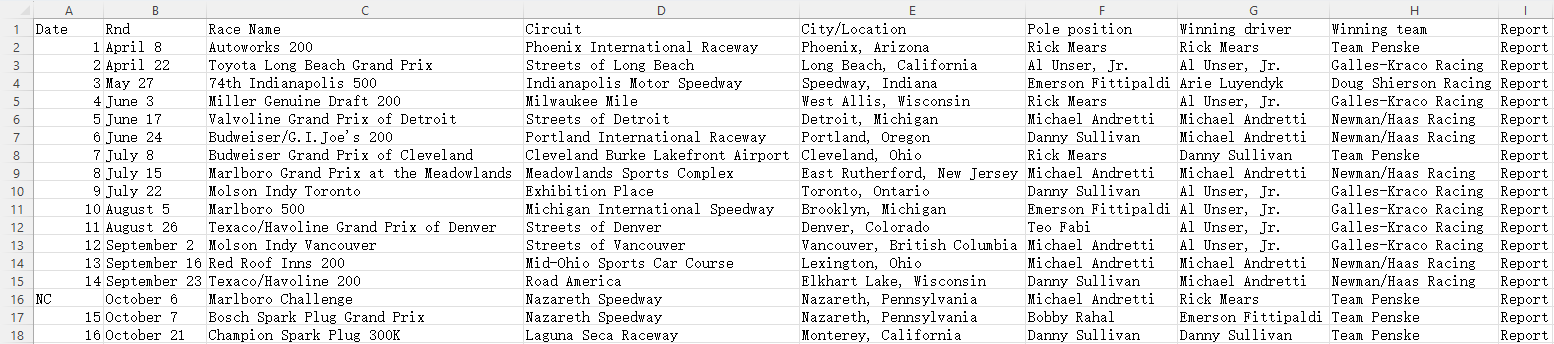}
\\
& \textbf{Groundtruth Answer:} 15\\

\midrule
SFT & \textbf{Think:} \textit{To determine the number of of winning States that represent,  we need to count the occurrences of U.S. drivers in the range B1:B10. This will give us the total count of cells that contain the value "A1".} \\
& \textbf{Generated Formula:} \texttt{=COUNTIF(B1:B10, A1)} \\\\
& \textbf{Execution Result:} 0 \\
\\
& \textbf{Execution Accuracy:} 0.0 \\

\midrule
RL & \textbf{Think:} \textit{1. To answer this question, I need to look into the table provided in the question. The question is asking to determine how many times a driver from the United States (U.S.) won a race based on a given table.
2. Since the question is focused on the winning drivers, I need to look for a column in the table that contains the names of the winning drivers.
3. Given that the question is asking to count how many times a U.S. driver won, I will need to look for a column that contains the names of the winning drivers and then count how many of those names are U.S. drivers.
4. Since the question is not providing a specific column or row in the table where the winning drivers' names are listed, I will need to assume that the winning drivers' names are listed in a column that I can reference in the table.} \\
& \textbf{Generated Formula:} \texttt{=COUNTIF(G2:G18, "Rick Mears") + COUNTIF(G2:G18, "Al Unser, Jr.") + COUNTIF(G2:G18, "Michael Andretti") + COUNTIF(G2:G18, "Danny Sullivan")} \\
\\
& \textbf{Execution Result:} 15 \\
\\
& \textbf{Execution Accuracy:} 1.0 \\
\bottomrule
\end{tabular}
\label{tab:example_rl}
\end{table*}

\clearpage
\section{Spreadsheet Formula Operators in Symbolic Table Reasoning}
\label{ap:operator}

Spreadsheet formulas are Turing complete~\citep{smalley2023excel}, but in practice, only a core subset of functions is commonly used. To guide the language model toward effective formula generation and support future research, we identify a set of representative operators in Table~\ref{tab:operator}. This table lists each operator’s name, symbol, definition, and a representative example—primarily covering basic arithmetic and aggregation operations. Acknowledging the model’s limited formula knowledge at the beginning of training, we explicitly introduce these operators during prompting, while still allowing the model to use any formula supported by our execution engine.

The formula operators can be viewed as the action space in reinforcement learning for symbolic table reasoning. The selected operators are designed to cover the majority of symbolic reasoning needs in table-based question answering, including row/column indexing, numerical aggregation, and conditional filtering. Focusing on a fixed set of operators facilitates interpretable error analysis and enables fine-grained tracking of formula usage patterns during both training and evaluation. This curated set also provides a natural foundation for curriculum learning strategies—starting with simpler operators and progressively introducing more complex ones, such as nested conditions and lookup functions.

\begin{table*}[ht]
  \centering
  \caption{Representative spreadsheet formula operators in symbolic table reasoning: Symbols, Definitions, and Examples.}
  \resizebox{\textwidth}{!}{%
    \begin{tabular}{llll}
      \toprule
      \textbf{Name} & \textbf{Symbol} & \textbf{Description} & \textbf{Example} \\
      \midrule
      PLUS                     & +      & Adds two numbers together                                         & \texttt{=A1 + A2}                   \\
      MINUS                    & --     & Subtracts one number from another                                 & \texttt{=A1 - A2}                   \\
      MULTIPLY                 & *      & Multiplies two numbers together                                   & \texttt{=A1 * A2}                   \\
      DIVIDE                   & /      & Divides one number by another                                     & \texttt{=A1 / A2}                   \\
      SUM                      & SUM    & Sums a range of numbers                                           & \texttt{=SUM(A1:A10)}               \\
      AVERAGE                  & AVERAGE& Calculates the average of a range of numbers                      & \texttt{=AVERAGE(A1:A10)}           \\
      COUNT                    & COUNT  & Counts the number of numbers in a range                           & \texttt{=COUNT(A1:A10)}             \\
      MAX                      & MAX    & Finds the maximum number in a range                                & \texttt{=MAX(A1:A10)}               \\
      MIN                      & MIN    & Finds the minimum number in a range                                & \texttt{=MIN(A1:A10)}               \\
      EQUAL                    & =      & Returns TRUE if the two values are equal                          & \texttt{=A1 = A2}                   \\
      NOT\_EQUAL               & <>     & Returns TRUE if the two values are not equal                      & \texttt{=A1 <> A2}                  \\
      GREATER\_THAN            & >      & Returns TRUE if the first value is greater than the second        & \texttt{=A1 > A2}                   \\
      LESS\_THAN               & <      & Returns TRUE if the first value is less than the second           & \texttt{=A1 < A2}                   \\
      GREATER\_THAN\_OR\_EQUAL & >=     & Returns TRUE if the first value is greater than or equal to second& \texttt{=A1 >= A2}                  \\
      LESS\_THAN\_OR\_EQUAL    & <=     & Returns TRUE if the first value is less than or equal to second   & \texttt{=A1 <= A2}                  \\
      AND                      & AND    & Returns TRUE if all arguments are TRUE                            & \texttt{=AND(A1, A2)}               \\
      OR                       & OR     & Returns TRUE if any argument is TRUE                              & \texttt{=OR(A1, A2)}                \\
      NOT                      & NOT    & Returns TRUE if the argument is FALSE                             & \texttt{=NOT(A1)}                   \\
      IF                       & IF     & Returns one value if a condition is TRUE and another if FALSE     & \texttt{=IF(A1 > 10, "Yes", "No")}   \\
      TRUE                     & TRUE   & Returns TRUE                                                      & \texttt{=TRUE}                      \\
      FALSE                    & FALSE  & Returns FALSE                                                     & \texttt{=FALSE}                     \\
      INDEX                    & INDEX  & Returns the value of a cell at a specific row and column          & \texttt{=INDEX(A1:A10, 1)}          \\
      MATCH                    & MATCH  & Returns the position of an item in an array (see syntax below)    & \texttt{=MATCH("value", A1:A10, 0)} \\
      \bottomrule
    \end{tabular}%
  }
  \label{tab:operator}
\end{table*}

\clearpage
\section{Prompts Used in the Experiments}
\label{ap:prompt}

\begin{figure*}[ht!]
    \centering
    \includegraphics[width=0.95\textwidth]{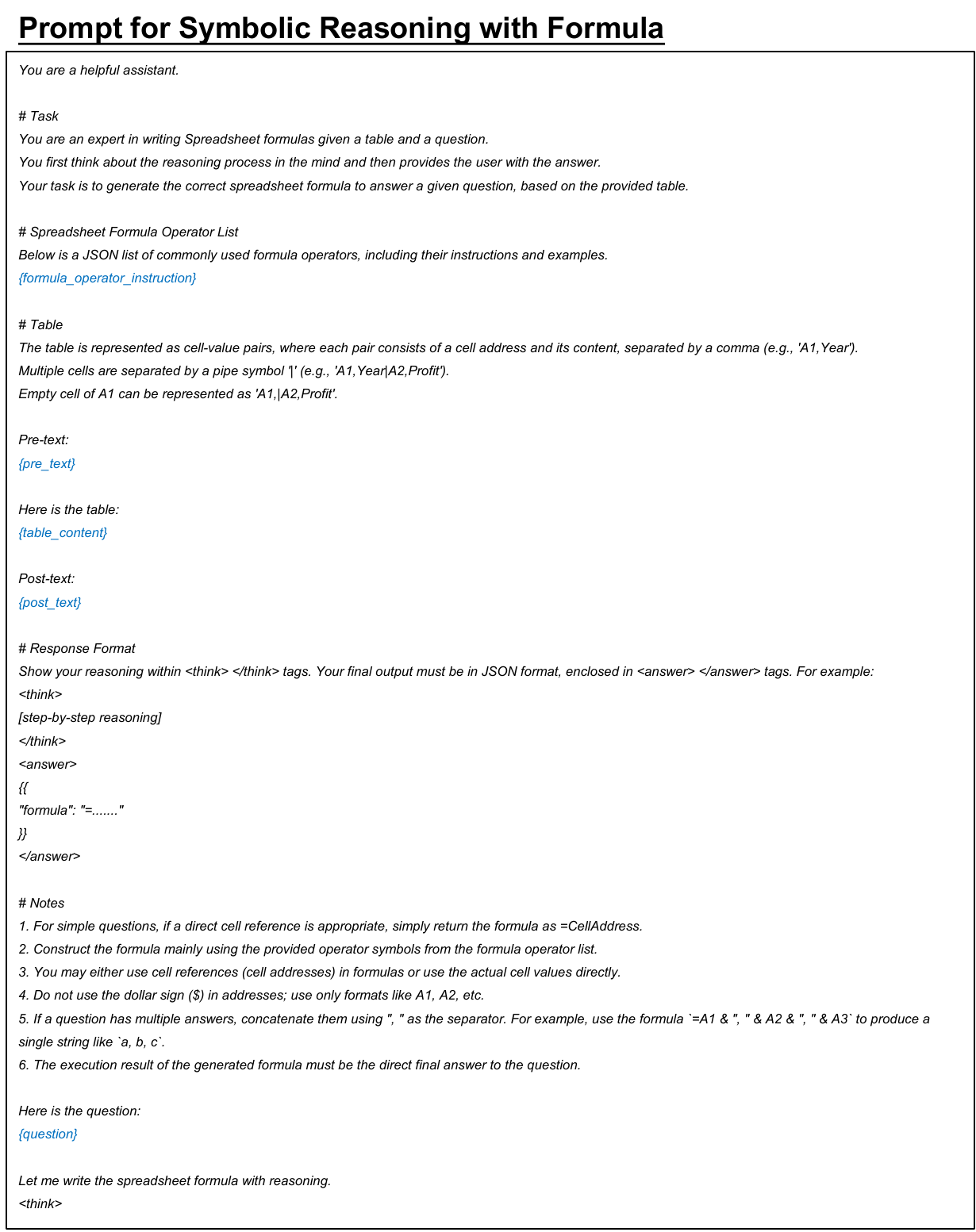}
    \caption{Prompt for symbolic reasoning with formula. Blue text indicates placeholders for variables within the prompt.}
    \label{fig:prompt_formula}
\end{figure*}

\begin{figure*}[ht!]
    \centering
    \includegraphics[width=0.95\textwidth]{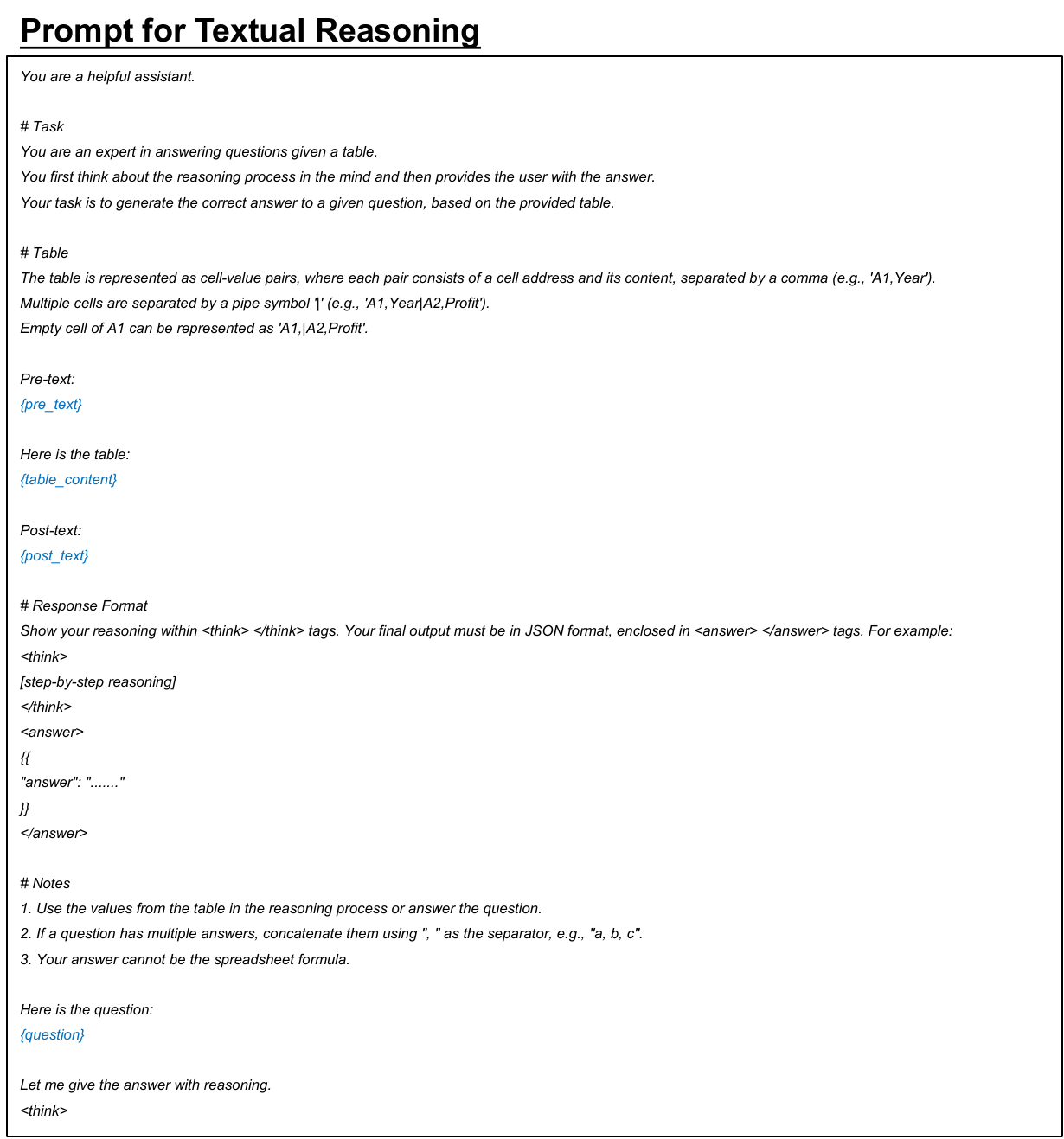}
    \caption{Prompt for textual reasoning. Blue text indicates placeholders for variables within the prompt.}
    \label{fig:prompt_text}
\end{figure*}

\begin{figure*}[ht!]
    \centering
    \includegraphics[width=0.95\textwidth]{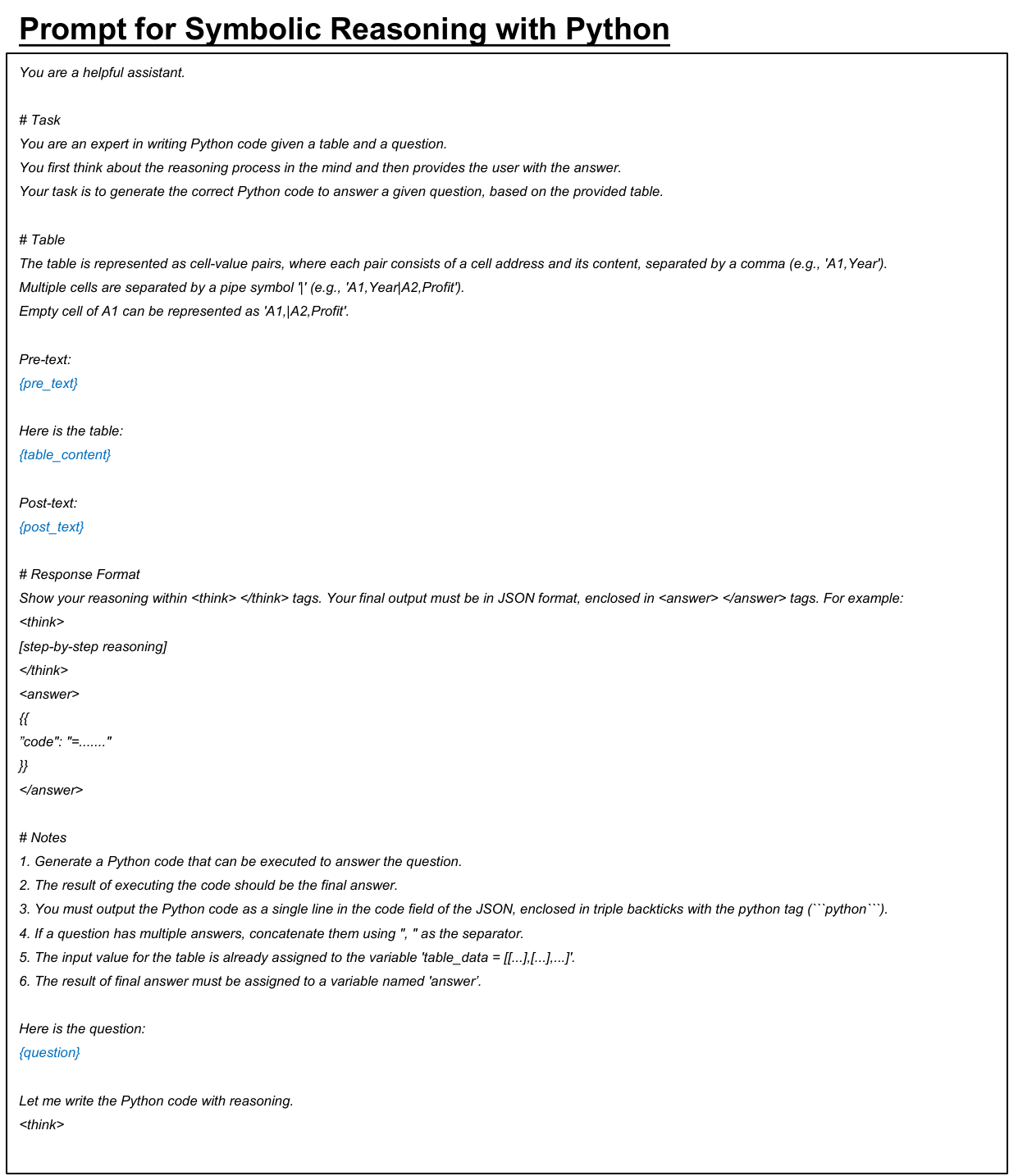}
    \caption{Prompt for symbolic reasoning with Python. Blue text indicates placeholders for variables within the prompt.}
    \label{fig:prompt_python}
\end{figure*}

\begin{figure*}[ht!]
    \centering
    \includegraphics[width=0.95\textwidth]{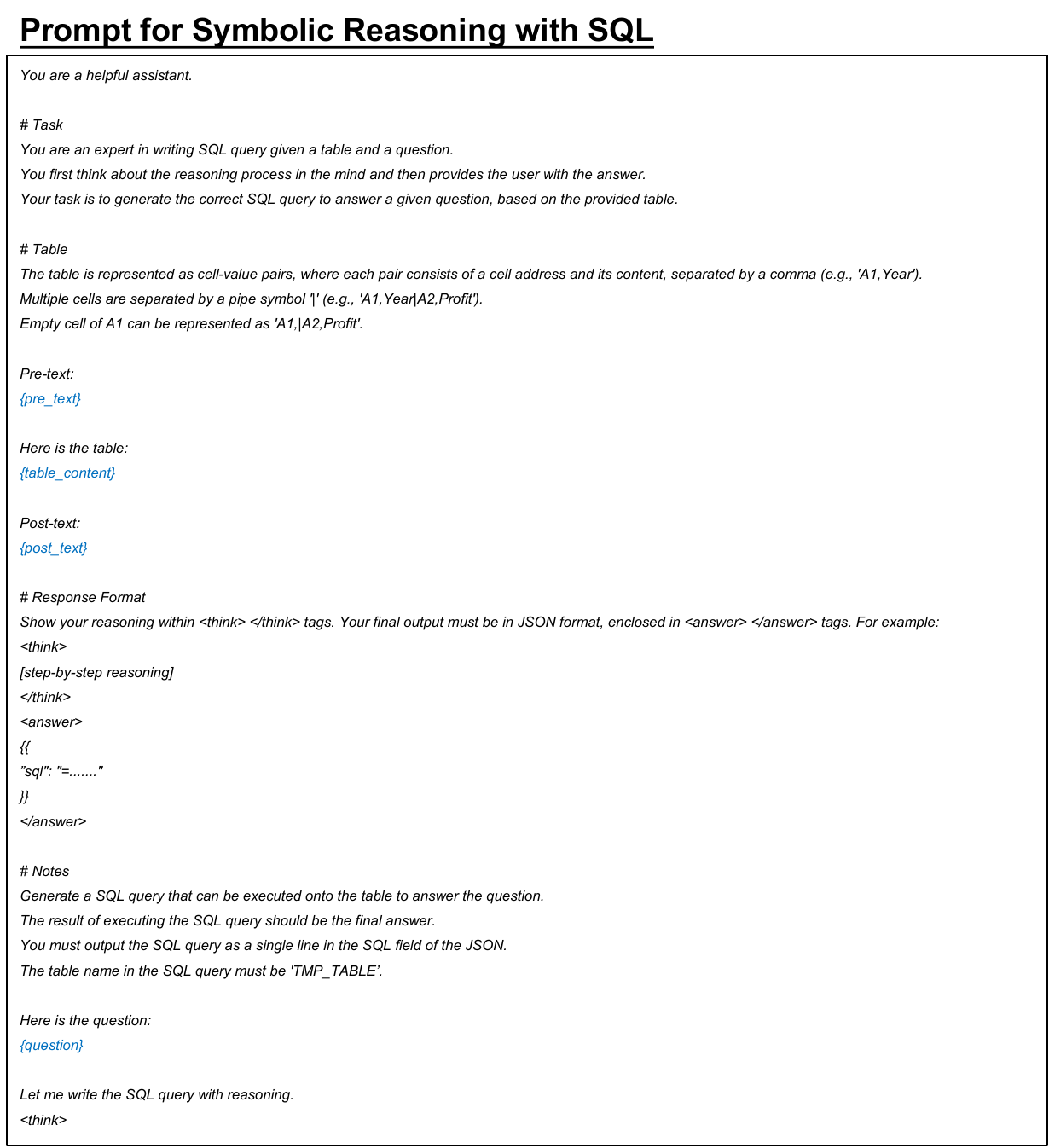}
    \caption{Prompt for symbolic reasoning with SQL. Blue text indicates placeholders for variables within the prompt.}
    \label{fig:prompt_sql}
\end{figure*}

\begin{figure*}[ht!]
    \centering
    \includegraphics[width=0.95\textwidth]{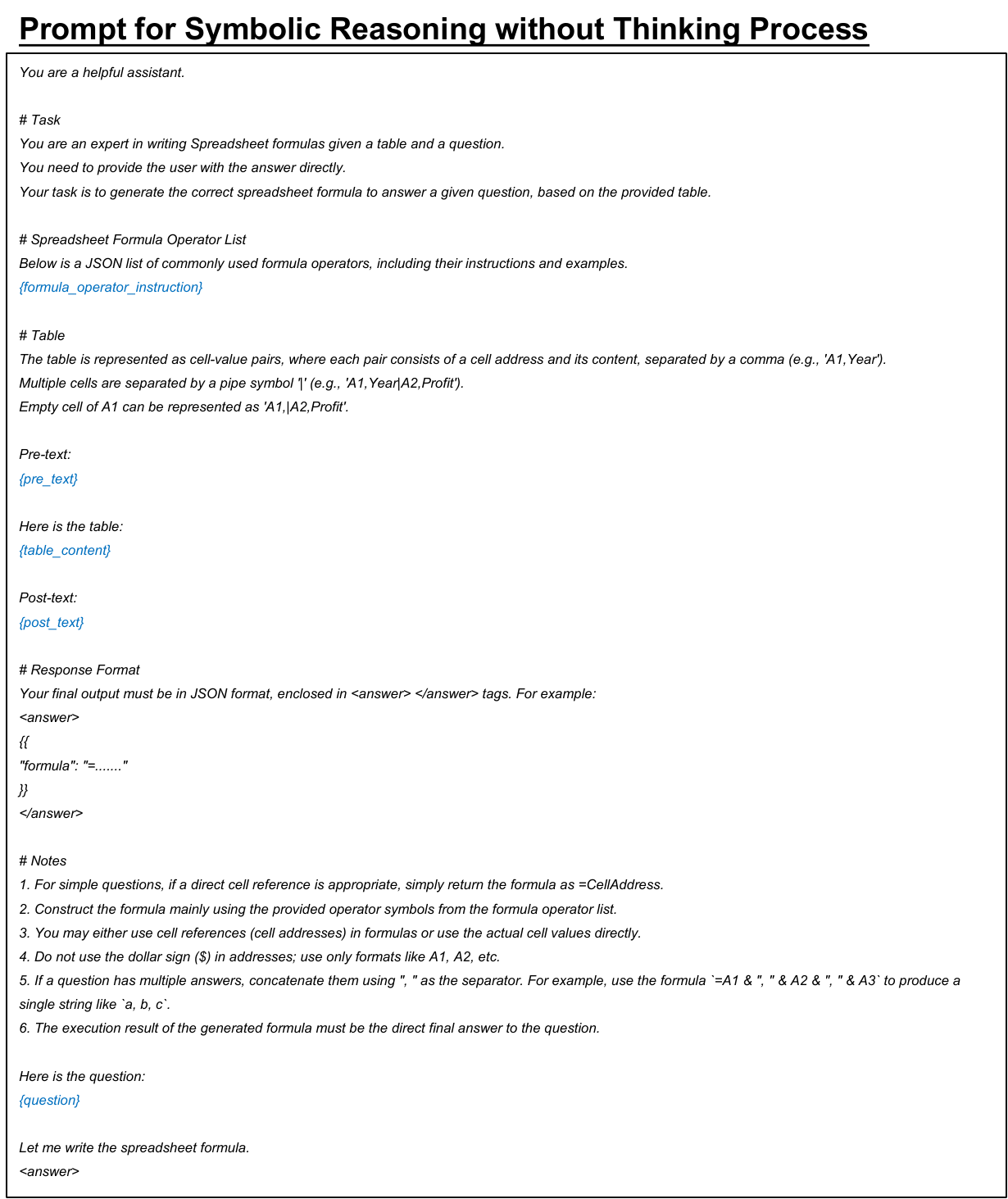}
    \caption{Prompt for symbolic reasoning with thinking process. Blue text indicates placeholders for variables within the prompt.}
    \label{fig:prompt_wo_reason}
\end{figure*}

Figures~\ref{fig:prompt_formula}, \ref{fig:prompt_text}, \ref{fig:prompt_python}, \ref{fig:prompt_sql}, and~\ref{fig:prompt_wo_reason} illustrate the prompts used in our experiments across zero-shot inference, supervised fine-tuning (SFT), reinforcement learning (RL), and evaluation.

\textit{Pre-text} and \textit{Post-text} refer to optional unstructured context surrounding the table, such as the data description format used in FinQA~\citep{chen-etal-2021-finqa}. \textit{formula operator instruction} represents the textual representations and usage guidelines of the representative spreadsheet formula operators in symbolic table reasoning, as detailed in Appendix~\ref{ap:operator}.

These prompts serve as examples rather than optimal templates. They may vary across tasks and can be further optimized for better performance.

% \clearpage
% \section{Notation Table}
% \label{ap:notation}

% Table~\ref{tab:notation} provides a comprehensive list of the notations used throughout this paper, along with their corresponding descriptions. This table serves as a quick reference to help readers better understand
% the concepts presented in our work.

% \input{tables/notation}

% \clearpage
% \section{The Use of Large Language Models (LLMs)}
% In this paper, we use an LLM (\textit{GPT-4o}) as the source for distilling the supervision data used in the cold-start SFT stage. In addition, LLMs were used to assist with manuscript writing and polishing.

%%%%%%%%%%%%%%%%%%%%%% Lang's LAST PAGE %%%%%%%%%%%%%%%%%%%%

\clearpage
\vspace*{\fill}
\begin{center}
This is the LAST PAGE.
2025.05.16
\end{center}
\vspace*{\fill}

\end{document}